\documentclass[11pt]{article}

\usepackage[final]{acl}

\usepackage{times}
\usepackage{latexsym}

\usepackage[T1]{fontenc}

\usepackage{fontawesome5}
\usepackage{booktabs}
\usepackage{array}
\usepackage{multirow}
\usepackage{caption}
\newcolumntype{L}[1]{>{\raggedright\arraybackslash}p{#1}}
\newcolumntype{C}[1]{>{\centering\arraybackslash}p{#1}}

\usepackage[most]{tcolorbox} 
\usepackage{enumitem}        
\usepackage{xcolor}          
\definecolor{summarybg}{RGB}{230, 240, 248}   
\definecolor{summarytitle}{RGB}{31, 119, 180} 
\definecolor{Ideas}{RGB}{251, 192, 45} 
\definecolor{Summary}{RGB}{44, 160, 44} 
\usepackage{fontawesome5} 
\newtcolorbox{summarybox}[1]{
    colback=summarybg,
    colframe=summarybg,
    arc=0mm,
    boxrule=0pt,
    left=2mm, right=2mm, top=2mm, bottom=2mm,
    fontupper=\scriptsize,
    title={\textbf{\scriptsize #1}},
    coltitle=summarytitle,
    attach title to upper,
    after title={\smallskip}
}

\usepackage[utf8]{inputenc}

\usepackage{tikz}
\usepackage{xcolor}
\usetikzlibrary{shapes.geometric, arrows.meta, positioning, calc, backgrounds}

\definecolor{memoryorange}{RGB}{160, 215, 180}
\definecolor{reflectiveyellow}{RGB}{160, 200, 240}
\definecolor{experiencegreen}{RGB}{245, 170, 170}

\definecolor{memoryorangelight}{RGB}{195, 230, 210}
\definecolor{reflectiveyellowlight}{RGB}{195, 220, 245}
\definecolor{experiencegreenlight}{RGB}{250, 205, 205}

\definecolor{memoryorangelighter}{RGB}{220, 242, 230}
\definecolor{reflectiveyellowlighter}{RGB}{220, 235, 250}
\definecolor{experiencegreenlighter}{RGB}{252, 225, 225}

\definecolor{memoryorangelightest}{RGB}{238, 250, 242}
\definecolor{reflectiveyellowlightest}{RGB}{238, 245, 252}
\definecolor{experiencegreenlightest}{RGB}{255, 245, 245}

\definecolor{arrowtop}{RGB}{160, 215, 180}
\definecolor{arrowmiddle}{RGB}{160, 200, 240}
\definecolor{arrowbottom}{RGB}{245, 170, 170}

\pgfdeclareverticalshading{triplestopgradient}{100bp}{
    color(10bp)=(arrowbottom);
    color(50bp)=(arrowmiddle);
    color(90bp)=(arrowtop)
}

\usepackage{microtype}

\usepackage{graphicx}
\usepackage{amsmath}

\title{From Storage to Experience: A Survey on the Evolution of LLM Agent Memory Mechanisms\thanks{Our continuously updated list of papers and resources is available at \url{https://github.com/FeishuLuo/Evolving-LLM-Agent-Memory-Survey}.}}

\author{Jinghao Luo$^2$\thanks{Equal contribution.}, 
        Yuchen Tian$^1$\footnotemark[2]\thanks{Project leader.}, 
        Chuxue Cao$^3$, Ziyang Luo$^1$, Hongzhan Lin$^1$,\\ 
        \textbf{Kaixin Li}$^4$, \textbf{Chuyi Kong}$^1$, \textbf{Ruichao Yang}$^5$, \textbf{Jing Ma}$^1$\thanks{Corresponding author.} \\ 
        $^1$Hong Kong Baptist University \,
        $^2$South China Normal University\\
        $^3$Hong Kong University of Science and Technology\\
        $^4$National University of Singapore \,
        $^5$University of Science and Technology Beijing\\
        \texttt{FeishuEcho@outlook.com, \{yctian, majing\}@comp.hkbu.edu.hk}}

\begin{document}
\maketitle
\begin{abstract}
Large Language Model (LLM)-based agents have fundamentally reshaped artificial intelligence by integrating external tools and planning capabilities.
%
While memory mechanisms have emerged as the architectural cornerstone of these systems, current research remains fragmented, oscillating between operating system engineering and cognitive science. This theoretical divide prevents a unified view of technological synthesis and a coherent evolutionary perspective.
To bridge this gap, this survey proposes a novel evolutionary framework for LLM agent memory mechanisms, formalizing the development process into three stages: \textbf{Storage} (trajectory preservation), \textbf{Reflection} (trajectory refinement), and \textbf{Experience} (trajectory abstraction). We first formally define these three stages before analyzing the three core drivers of this evolution: the necessity for long-range consistency, the challenges in dynamic environments, and the ultimate goal of continual learning. Furthermore, we specifically explore two transformative mechanisms in the frontier Experience stage: active exploration and cross-trajectory abstraction. By synthesizing these disparate views, this work offers robust design principles and a clear roadmap for the development of next-generation LLM agents.

\end{abstract}

\section{Introduction}\label{Introduction}
In recent years, the rapid advancement of Large Language Models (LLMs) has fundamentally reshaped the landscape of artificial intelligence~\citep{llama,gpt4o,qwen3}. To augment the capabilities of LLMs, researchers have developed LLM-based agents that integrate LLMs with external tools and modular components, thereby enabling planning, tool use, and environmental interaction~\citep{Yao2022ReActSR,toollm,MCPU}. However, the inherent statelessness of LLMs poses a critical challenge: it hinders agents from maintaining logical consistency across complex, multi-step tasks and precludes learning from prior interactions, often resulting in recurring reasoning errors~\citep{Huang2023LargeLM, Xiong2025HowMM, cao2026diffcot}. Consequently, the development of effective memory mechanisms has emerged as an architectural cornerstone. By mitigating this deficiency, memory mechanisms underpin the robust operation of LLM-based agents and pave the way for self-evolution~\cite{Wang2023ASO, Packer2023MemGPTTL, Wu2025EvolveRSL}.

\begin{figure*}[!h]
    \centering
    \includegraphics[width=1\linewidth]{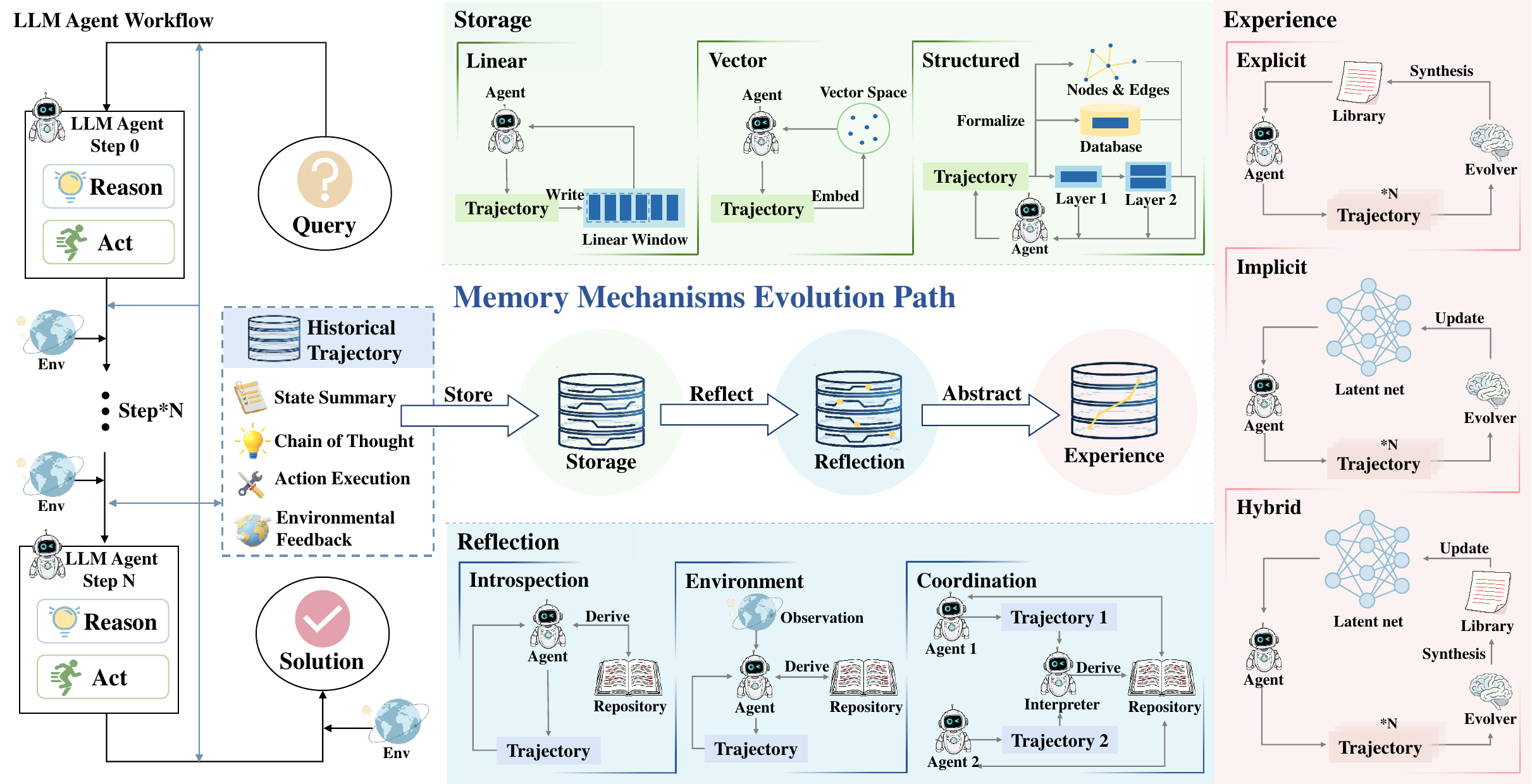}
    \vspace{-0.3cm}
    \caption{Overview of the LLM agent memory mechanisms.}
    \label{fig:overview}
    \vspace{-0.3cm}
\end{figure*}

We identify two primary obstacles to advancing memory mechanisms for LLM agents: (i) \textit{Paradigmatic Fragmentation}: Existing methodologies oscillate between two weakly integrated paradigms. One focuses on engineering, adopting design principles from operating systems for the management of memory data~\citep{Packer2023MemGPTTL, Hu2024HiAgentHW, Kang2025MemoryOO}, while the other draws inspiration from cognitive science and psychology to simulate mechanisms for the formation, consolidation, and retrieval of human memory~\citep{Zhong2023MemoryBankEL, Hou2024MyAU, Xu2025AMEMAM}. This lack of synergistic progress results in a fragmented body of research, preventing the formation of a coherent and continuous trajectory of evolution. (ii) \textit{The Absence of Technological Synthesis}: Although numerous methods address isolated stages of memory processing, the field lacks a cohesive summary of the critical technologies that have historically propelled memory mechanism advancement~\citep{Xu2025AMEMAM, Yang2025CoarsetoFineGM, Zhang2025MemEngineAU}. Existing surveys have not sufficiently isolated these key technical drivers from general methodologies~\cite{Wu2025FromHM, Du2025RethinkingMI, doi:10.36227/techrxiv.176539617.79044553/v1, Cao2025RememberMR}. Consequently, the core technologies remain obscure, leaving future researchers without a clear roadmap of which innovations are robust enough to build upon.

While recent surveys have examined memory mechanisms for LLM agent systems, they lack a unified evolutionary perspective. This limitation obscures the internal drivers of memory development and impedes the in-depth exploration of architectures for next-generation agents. Specifically, \citet{Zhang2024ASO} focuses on the classification of engineering modules, but fails to systematically expound on the logic behind critical technological transformations throughout their development. Furthermore, while \citet{Hu2025MemoryIT} addresses the dynamic processes of memory, its perspective remains confined to static functional categorizations, failing to reveal the underlying principles of dynamic evolution inherent to memory mechanisms.

To address these limitations, we propose a framework for memory mechanisms in LLM-based agents centered on dynamic evolution. We formalize this evolutionary process into three distinct stages: (i) \textbf{Storage}, which constructs diverse storage modes focused on the faithful recording of historical interaction trajectories; (ii) \textbf{Reflection}, which introduces a loop for dynamic evaluation to actively manage and refine these records; and (iii) \textbf{Experience}, which implements prospective guidance by abstracting high-level behavior patterns and strategies from clustered interactions (\S\ref{Background}).

Building upon the proposed three stages of memory mechanisms, this survey follows a "Why-How-What" logic to address three interconnected research questions:~\textbf{RQ1: Why do memory mechanisms evolve?} reveals how the requirements for long-range consistency, dynamic environment interaction, and continual learning serve as core catalysts driving mechanistic evolution (\S\ref{Evolutionary Drivers});~\textbf{RQ2: How do memory mechanisms evolve?} delineates the evolutionary path from Storage to Reflection and then to Experience, analyzing the fundamental structural shifts involved (\S\ref{Evolutionary Path}); and~\textbf{RQ3: What changes does Experience bring?} provides an in-depth analysis of how frontier paradigms in the Experience stage, such as proactive exploration and cross-trajectory abstraction, address the bottlenecks in agent adaptability and autonomy (\S\ref{Transformative Experience}).


Finally, we outline future directions for LLM agent memory mechanisms. First, we emphasize that memory mechanisms should adopt more dynamic triggering modes based on task types (\S\ref{Future Direction}). Second, we highlight that the construction of working memory is a vital core of memory mechanisms. Next, we advocate for the development of more comprehensive datasets for memory mechanisms, especially for the Experience stage. Finally, we establish the coordination of distributed shared memory and the fusion of multimodal memory as critical breakthroughs for future research.

The overview of this survey and related datasets is documented in Appendix~\S\ref{Overview} and~\S\ref{Datasets}, respectively.

\section{Background}\label{Background}
\subsection{The LLM Agent Framework}
We formalize an LLM-based agent as a decision-making entity parameterized by $\theta$, interacting with a dynamic environment $\mathcal{E}$. The agent's operation is governed by a policy $\pi_\theta$, which maps the current context to a probability distribution over the action space $\mathcal{A}$.

At time step $t$, the agent receives an observation $o_t \in \mathcal{O}$ and retrieves relevant information $m_t$ from its memory module $\mathcal{M}$. The generated action $a_t$ is sampled as follows:
{\setlength{\abovedisplayskip}{0.1cm}
\setlength{\belowdisplayskip}{0.1cm}
\begin{equation}
    a_t \sim \pi_\theta(a_t \mid \mathcal{I}, o_t, m_t),
\end{equation}}
where $\mathcal{I}$ denotes the static system instruction, and $m_t = \text{Retrieve}(\mathcal{M}, o_t)$ represents the context-specific memory. Crucially, we distinguish between the global memory repository $\mathcal{M}$ and its retrieved instantiation $m_t$ at time $t$. In this survey, we define ``LLM agent memory'' $\mathcal{M}$ as an externalized repository that bridges the frozen parametric knowledge in $\theta$ and the evolving environmental dynamics. 

\subsection{Taxonomy}
We classify the evolution of memory mechanisms into three tiers based on the level of information abstraction and cognitive processing.

\noindent\textbf{Storage.}
Storage serves as the foundational layer. Unlike higher-level mechanisms, storage preserves trajectories with minimal transformation, maintaining a one-to-one correspondence between memory entries and execution traces. We define a trajectory $\tau$ as a chronological sequence of observation-action pairs within a task session:
{\setlength{\abovedisplayskip}{0.1cm}
\setlength{\belowdisplayskip}{0.1cm}
\begin{equation}
    \tau = \langle (o_1, a_1), \dots, (o_T, a_T) \rangle.
\end{equation}}

The raw storage $\mathcal{M}_{raw}$ is formally defined as a cumulative set of historical trajectories:
{\setlength{\abovedisplayskip}{0.1cm}
\setlength{\belowdisplayskip}{0.1cm}
\begin{equation}
    \mathcal{M}_{raw} = \{ \tau_i \}_{i=1}^N, \quad \tau_i \in \mathcal{T},
\end{equation}}
where $\mathcal{T}$ represents the space of all possible interaction trajectories.

\noindent\textbf{Reflection.}
Reflection is modeled as a semantic transformation mapping $\mathcal{F}_{ref}: \mathcal{T} \to \mathcal{S}$, where $\mathcal{S}$ denotes the space of evaluated or corrected reasoning paths. Similar to the storage phase, Reflection functions as a mechanism to populate the global repository $\mathcal{M}$, but with a focus on quality density rather than raw fidelity.

It operates by analyzing a completed trajectory $\tau_i$ to generate a refined memory unit $m'_i$, which encapsulates critiques or corrective insights:
{\setlength{\abovedisplayskip}{0.1cm}
\setlength{\belowdisplayskip}{0.1cm}
\begin{equation}
    m'_i = \mathcal{F}_{ref}(\tau_i \mid \phi),
\end{equation}}
where $\phi$ represents the evaluation criteria. The key distinction lies in the storage protocol: while standard Storage preserves raw interaction logs, Reflection acts as a semantic filter, injecting processed insights back into the repository ($\mathcal{M} \leftarrow \mathcal{M} \cup \{ m'_i \}$). Once stored, $m'_i$ becomes an independent memory entry, decoupling the valuable logic from the specific noise of the original trajectory $\tau_i$ and serving as a refined reference for future retrieval.

\noindent\textbf{Experience.}
Experience represents the highest cognitive layer, characterized by cross-trajectory abstraction. This stage aims to satisfy the Minimum Description Length (MDL) principle by compressing redundant trajectories into generalized schemas.
Let $\mathcal{T}_{batch} \subset \mathcal{M}_{raw}$ be a subset of topologically similar trajectories. We define the Experience function $\mathcal{F}_{exp}$ as an inductive operator that extracts a set of universally applicable rules $\mathcal{K}$:
{\setlength{\abovedisplayskip}{0.1cm}
\setlength{\belowdisplayskip}{0.1cm}
\begin{equation}
    \mathcal{T}_{batch} = \{ \tau_i \mid \text{Sim}(\tau_i, \tau_j) > \epsilon \},
\end{equation}}
{\setlength{\abovedisplayskip}{0.1cm}
\setlength{\belowdisplayskip}{0.1cm}
\begin{equation}
    \mathcal{K} = \mathcal{F}_{exp}(\mathcal{T}_{batch}) \quad \text{s.t.} \quad |\mathcal{K}| \ll \sum_{\tau \in \mathcal{T}_{batch}} |\tau|.
\end{equation}}
Formally, $\mathcal{K}$ serves as a policy prior that elevates $\pi_\theta$ beyond 
rule consistent actions, enabling decision-making at a higher level of abstraction.

{\setlength{\abovedisplayskip}{0.1cm}
\setlength{\belowdisplayskip}{0.1cm}
\begin{summarybox}{Summary \& Ideas - Defining the Stages of memory mechanisms}
    \begin{itemize}[leftmargin=1.8em, nosep] 
        \item[{\makebox[1em][c]{\textcolor{Summary}{\faBook}}}] Defining the Stages of memory mechanisms We provide a formal definition for the workflow of actions in LLM agents and characterize the memory mechanisms within this process as three stages of evolution: Storage, Reflection, and Experience.   (\textit{cf}. Figure~\ref{fig:overview})
        \item[{\makebox[1em]{\textcolor{Ideas}{\faLightbulb}}}] We define these mechanisms based on the degree of utilization of memory, without introducing other related technical definitions.
        \item[{\makebox[1em]{\textcolor{Ideas}{\faLightbulb}}}] The three stages of evolution do not represent complete substitution but are framed within an evolutionary perspective; a mechanism for memory may still retain certain characteristics of the preceding stage, even as its essence has transitioned into the subsequent stage.
    \end{itemize}
\end{summarybox}}

\section{Evolutionary Drivers}\label{Evolutionary Drivers}

To facilitate a comprehensive understanding regarding the evolution of memory mechanisms for LLM agents, we first address the fundamental question \textbf{RQ1: Why do memory mechanisms evolve?} In this section, we examine three core requirements for LLM agents to investigate how they drive the progression of memory mechanisms, thereby bridging the gap between models from pretraining and the real world.

\subsection{Long-Term Consistency}

Consistency across long horizons constitutes a prerequisite for the deployment of LLM agents within the real world and serves as the primary impetus for the early evolution of memory mechanisms. Although large language models exhibit robust local coherence within the context window, they frequently encounter issues such as redundant exploration, accumulation of errors, and discontinuities in reasoning during interactions involving multiple steps. We analyze the necessity of consistency over long durations through two dimensions: consistency of state and consistency of goals.


\noindent\textbf{Consistency of State.}
The inherent statelessness of LLM agents results in a deficiency of internal mechanisms for explicit anchoring, which has catalyzed the emergence of modules for memory~\citep{Huang2023AdvancingTA, Sumers2023CognitiveAF,Packer2023MemGPTTL}. First, these modules maintain internal states for reasoning to ensure the coherence of thought~\citep{Yao2023TreeOT, Sun2025ScalingLL}; second, they synchronize the cognition of the agent with the external world to prevent erroneous decisions arising from inaccurate internal perceptions~\citep{Majumder2023CLINAC, Yang2025CoarsetoFineGM}; finally, they internalize interactions into persistent traits of the persona to ensure uniformity in behavior~\citep{Park2023GenerativeAI, Westhuer2025EnablingPL, Liang2025AIMB}.


\noindent\textbf{Consistency of Goals.}
Due to the inherent nature of planning by the agent, LLM agents frequently optimize for actions with local consistency, which results in a departure from objectives at the global level~\citep{Huang2024UnderstandingTP, Everitt2025EvaluatingTG}. Memory mechanisms mitigate this drift by providing persistent and explicit goals at a high level~\citep{Hu2024HiAgentHW, Li2025HiPlanHP}. Furthermore, in systems with multiple agents, shared memory regarding goals can transform isolated behaviors into coordinated execution by the collective, thereby maintaining the unity of the final objective~\citep{Gao2024AgentScopeAF, Liu2025RCRRouterER}.


\subsection{Dynamic Environments}
The dynamic characteristics of the environment constitute a more enduring impetus for the evolution of memory mechanisms. In contrast to static benchmarks, the interplay between temporal validity and causality in real-world settings renders fixed patterns for reasoning and static forms of storage rapidly fragile.

\begin{figure}[!h]  
    \centering
    \includegraphics[width=1\linewidth]{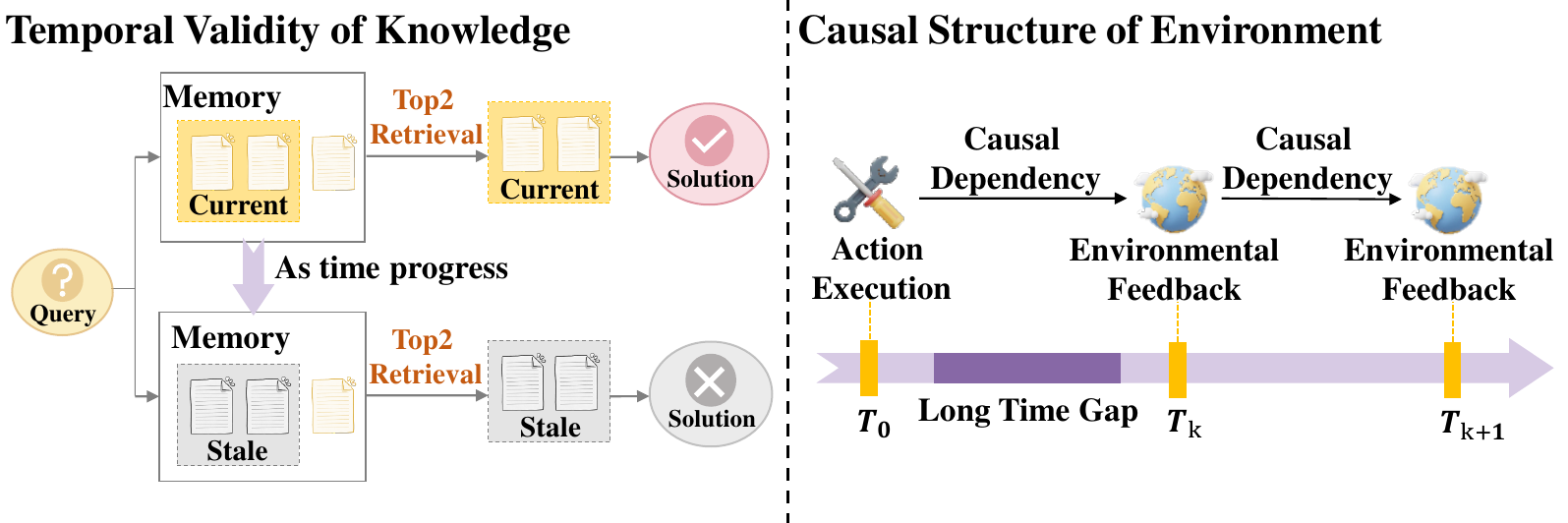}
    \caption{The Drivers in Dynamic Environments.}
    \label{fig:2}
\end{figure}

\noindent\textbf{The Temporal Validity of Knowledge.}
Knowledge within environments of a dynamic nature is typically conditional rather than eternally valid~\citep{Lazaridou2021MindTG, Jang2022TemporalWikiAL, Ko2024GrowOVERHC}. As the environment progresses, strategies for action that were once correct may experience a gradual loss of utility. Crucially, knowledge that is outdated often fails without overt indication~\citep{luu2022timewaitsoneanalysis, Kalai2023CalibratedLM,  kasai2024realtimeqawhatsanswer}; although factually incorrect, such information may still exhibit significant relevance in its semantic representation. This necessity propels the evolution of memory mechanisms from the paradigm of static storage toward that of active management, integrating awareness of temporal factors, policies for decay, and methods for retrieval with enhanced flexibility~\citep{Zhong2023MemoryBankEL, Siyue2024MRAGAM, Salama2025MemInsightAM, Du2025MemR3MR, Houichime2025MemoryAR}.

\noindent\textbf{The Causal Structure of the Environment.}
Causal relationships within the complex real world involve delayed outcomes and cascading effects~\citep{Joshi2024COLDCR, Cui2025UncertaintyIC, Liu2025LargeLM}. This necessitates that memory mechanisms transcend the mere documentation of interactions to construct dependencies for causality of a complex nature across steps in time~\citep{Majumder2023CLINAC, Du2025MemoryT1RL, Raman2025REMIAN}. Consequently, planning with robustness is achieved through the realization of internal worlds characterized by consistency in causality~\citep{Tang2024WorldCoderAM, Kim2025CoExC, Bohnet2025EnhancingLP}.

\subsection{Continual Learning}
Continual learning represents the ultimate requirement for LLM agents. Deployment within an open world inevitably involves encountering patterns that reside outside of the distribution of training. Without the effective internalization of these memories into actionable knowledge for reuse, the LLM agent will remain confined to repetitive cycles of trial and error. Therefore, memory mechanisms must not only enable the reproduction of historical trajectories but also address the bottlenecks of scaling and the requirements for abstraction inherent in dense memory.

\noindent\textbf{Constraints on The Storage of Memory.}
Interaction with the real world over extended durations results in the linear expansion of memory in storage~\citep{Hu2023ChatDBAL, Packer2023MemGPTTL}. Early memory mechanisms utilized techniques such as vectorization to scale storage capacity. However, recent research indicates that the unrestricted expansion of memory is detrimental to the performance of LLM agents, as errors propagate within the system for memory and contaminate the efficacy of learning~\citep{Xiong2025HowMM, srivastava2025memorygraftpersistentcompromisellm}. This necessitates the exploration of more strategic policies for the addition and deletion of information within memory mechanisms~\cite{Du2025RethinkingMI, Liu2025ContextAA}.

\noindent\textbf{The Requirement for Experience.}
The memory of most LLM agents is of an episodic nature and remains restricted to specific tasks~\citep{Shinn2023ReflexionLA, Wang2023ASO}. This limitation necessitates the transformation of raw clusters of memory into experience to provide guidance for behavior across future scenarios. Consequently, research on memory mechanisms has begun to explore various methodologies for the abstraction of experience~\citep{Tang2025AgentKL, Cai2025FLEXCA, Xia2025Agent0US, Alakuijala2025MementoNM, Guo2025GenEnvDC}.

%
{\setlength{\abovedisplayskip}{0.1cm}
\setlength{\belowdisplayskip}{0.1cm}
\begin{summarybox}{Summary \& Ideas - Evolutionary Drivers}
    \begin{itemize}[leftmargin=1.8em, nosep] 
        \item[{\makebox[1em][c]{\textcolor{Summary}{\faBook}}}] Consistency over long durations, dynamic environments, and the capacity for continual learning drive the evolution of LLM agents from static models into autonomous systems capable of operation within the real world over extended horizons.
        \item[{\makebox[1em]{\textcolor{Ideas}{\faLightbulb}}}] The driving force for the evolution of memory mechanisms stems from a leap in research vision: from maintaining logical consistency in complex reasoning, to dynamic adaptation in environmental perception, and finally to the autonomous evolution of LLM agents.
    \end{itemize}
\end{summarybox}}
\section{Evolutionary Path}\label{Evolutionary Path}
Building upon these evolutionary drivers, we conduct an investigation in-depth into \textbf{RQ2: How do memory mechanisms evolve?} We categorize the trajectory of evolution into three primary stages: storage, reflection, and experience.
\subsection{Storage}
The stage of storage serves as the starting point for memory mechanisms, where the primary objective is to resolve the contradiction between the limited window of context within Large Language Models and the continuously expanding history of interaction. Memory mechanisms during this phase are dedicated to the faithful preservation of interaction trajectories $\tau_i$ to the greatest extent possible to maintain consistency in the actions of the agent.

\noindent\textbf{Linear.}
Linear storage represents the most direct method of recording, in which interaction trajectories are treated as a stream of tokens ordered by time and managed typically through a strategy of First-In, First-Out (FIFO). Research focuses on the extension of the window of context via modifications to the mechanism of attention or the encoding of position~\citep{Ratner2022ParallelCW, Xiao2023EfficientSL,Jin2024LLMML}, as well as the achievement of information sparsification through the mechanical reduction of noise~\citep{Zhang2023H2OHO, Jiang2023LLMLinguaCP,Xiao2024InfLLMTL}.

\noindent\textbf{Vector.}
Vector storage encodes interaction trajectories into a high-dimensional space, which greatly expands the capacity for the storage of memory. Such methods shift the focus of research from the design of storage toward the optimization of retrieval, including retrieval based on semantic proximity~\citep{Melz2023EnhancingLI, Liu2024MemLongMR, Das2024LarimarLL} as well as weighted retrieval that incorporates temporal decay and scores for importance~\citep{Zhong2023MemoryBankEL, Park2023GenerativeAI}.

\noindent\textbf{Structured.}
Structured storage employs explicit data architectures to transcend the limitations on capacity inherent in linear storage and the ambiguity associated with vector retrieval. For instance, these methods utilize the tabular formats of relational databases for the storage of memory~\citep{Hu2023ChatDBAL, Xue2023DBGPTED, Lee2025TrainingAT}, partition memory into distinct hierarchies to address the trade-off between storage capacity and speed of retrieval~\citep{Packer2023MemGPTTL, Lu2023MemoChatTL}, and directly model the history of interaction as a topological network of entities and relations~\citep{Modarressi2024MemLLMFL, Li2024GraphReaderBG}.

\subsection{Reflection}
Mechanisms for storage fail to address the quality of memory, as raw trajectories are inevitably contaminated by hallucinations, errors in logic, and ineffective attempts. This limitation necessitates a transition of memory mechanisms toward reflection. In this phase, memory is transformed from a passive recorder into an active critic, utilizing various signals of feedback to perform correction and denoising of past trajectories to enhance the quality of the repository of memory.

\noindent\textbf{Introspection.}
Introspective reflection conceptualizes the LLM agent as an autonomous critic that leverages the internal knowledge of the model to refine memory without the requirement for external feedback. Research in this area focuses on the correction of errors within trajectories~\citep{Liu2023ThinkinMemoryRA, Zhang2025ConstructingCS, Bohnet2025EnhancingLP, cao2026pushing}, the maintenance of the lifecycle of memory~\citep{Li2025CAMAC, Kang2025MemoryOO, Chhikara2025Mem0BP}, and the compression and distillation of long trajectories~\cite{Huang2025MemOrbAP, Han2025LEGOMemMP, Yang2025CoarsetoFineGM, Ye2025AgentFoldLW}.

\noindent\textbf{Environment.}
Environmental reflection treats signals from the external environment as the primary anchors for the reflection of memory to mitigate the issue of hallucinations. This approach focuses on the utilization of outcomes from the real world to proactively optimize policies for behavior~\citep{Sun2024EnhancingAL, Yan2025MemoryDrivenSF, Yan2025MemoryR1EL} and calibrate internal models of the world~\citep{Sun2024EnhancingAL, Xiao2025ToolMemEM, Sun2025PreferenceAwareMU}.

\noindent\textbf{Coordination.}
Collaborative reflection extends this process to the collective, leveraging the division of roles and consensus to overcome bottlenecks in the cognition of individuals. This mechanism facilitates the reflection of memory through the construction of societies of heterogeneous agents~\cite{Bo2024ReflectiveMC, Balestri2025NarrativeMI, Wang2025GraphCogentML, Ozer2025MARMultiAgentRI}.

\subsection{Experience}
Although reflection effectively mitigates noise and hallucinations, reflected memories are frequently fragmented and exhibit a high degree of dependence on context. This results in significant costs for retrieval and a heavy burden of inference for memory mechanisms when addressing new tasks. Moreover, recent research indicates that LLM agents often demonstrate a pronounced tendency to follow successful trajectories; corrected trajectories devoid of abstraction may still induce errors resulting from minor shifts in context. Consequently, memory in the stage of experience extracts universal heuristic wisdom by isolating similar trajectories from their specific contexts. This approach compresses the originally vast repository of memory and enables generalization to unknown environments through a form of intuition similar to that of humans. A detailed comparison between reflection and experience is summarized in Table~\ref{tab:reflection_vs_experience}.



\noindent\textbf{Explicit.}
Explicit experience represents the integration of symbols, extracting human-readable and editable experiences with a high level of generalizability from clusters of interaction trajectories. This allows the LLM agent to achieve a highly interpretable process of self-evolution by either concretizing experiences into natural language policies~\citep{Cai2025FLEXCA, Zhang2025GMemoryTH, Hassell2025LearningFS, Wan2025LoongFlowDE} or directly abstracting them into executable entities~\citep{Wang2025InducingPS, Zhang2025AccelOptAS, Shi2025YoutuAgentSA}.A further line distills accumulated trajectories into an evolvable skill library, coupling procedural abstractions with a lifecycle of induction, reuse, and refinement~\citep{Zhang2026MemSkillLA, Ni2026Trace2SkillDT}.

\noindent\textbf{Implicit.}
Implicit experience internalizes interaction histories into model parameters, aiming to resolve the inference overhead and context limitations inherent in explicit memory. Implicit experience can be realized by directly converting experiences into the model's intrinsic capabilities through fine-tuning~\citep{Alakuijala2025MementoNM, Zhai2025AgentEvolverTE, Zhang2025AgentLV, Tandon2025EndtoEndTT, Yu2025GuidedSL}. Furthermore, the research community is exploring the transformation of experience into latent variables within the model's hidden layers, which are then dynamically invoked during the inference process~\citep{Zhang2025LatentEvolveST, Zhang2025MemGenWG}.

\noindent\textbf{Hybrid.}
Hybrid experience establishes a dynamic cycle of accumulation and internalization. Through a mechanism for experience transfer, explicit experience is treated as a high-capacity cache, which is subsequently compressed and internalized into the implicit weights of the model through periodic updates of parameters~\citep{Wu2025EvolveRSL, liu2026exploratorymemoryaugmentedllmagent, Ouyang2025ReasoningBankSA, Xia2026SkillRLEA}.

{\setlength{\abovedisplayskip}{0.1cm}
\setlength{\belowdisplayskip}{0.1cm}
\begin{summarybox}{Summary \& Ideas - Evolutionary Path}
    \begin{itemize}[leftmargin=1.8em, nosep] 
        \item[{\makebox[1em][c]{\textcolor{Summary}{\faBook}}}] This chapter traces how LLM agent memory mechanisms evolve from faithful recording in Storage, to error correction in Reflection, and finally to high-order abstraction in Experience.
        \item[{\makebox[1em]{\textcolor{Ideas}{\faLightbulb}}}] Although a portion of the literature adopts the terminology of experience, these works may not fall within the conceptual scope of experience established for this survey.
        \item[{\makebox[1em]{\textcolor{Ideas}{\faLightbulb}}}] Due to space constraints, we do not provide detailed descriptions of these works within the evolutionary path; instead, some representative studies are presented in the Appendix~\S\ref{Detail within the Evolutionary Path}.
    \end{itemize}
\end{summarybox}}

\begin{table*}[t]
\centering
\small
\renewcommand{\arraystretch}{1.3}
\begin{tabular}{p{2.8cm}p{6.5cm}p{6cm}}
\toprule
\textbf{Dimension} & \textbf{Reflection} & \textbf{Experience} \\
\midrule
Functional signature & 
Intra-trajectory transformation: $\mathcal{F}_{\text{ref}}(\tau_i \mid \phi) = m'_i$ & 
Inter-trajectory induction: $\mathcal{F}_{\text{exp}}(\mathcal{T}_{\text{batch}}) = \mathcal{K}$ \\
Output form & 
Refined memory unit $m'_i$ tied to the original task context & 
Universal rules or skills $\mathcal{K}$ detached from any specific scenario \\
Retrieval dependency & 
Retrieved to assist semantically similar past tasks at inference time & 
Applicable to unseen scenarios as a policy prior, without trajectory-level matching \\
Representative works & 
Reflexion~\citep{Shinn2023ReflexionLA}, CLIN~\citep{Majumder2023CLINAC}, AgentFold~\citep{Ye2025AgentFoldLW}& 
FLEX~\citep{Cai2025FLEXCA}, MemSkill~\citep{Zhang2026MemSkillLA}, SkillRL~\citep{Xia2026SkillRLEA} \\
\bottomrule
\end{tabular}
\caption{Structural comparison between Reflection and Experience. While Reflection injects refined units $m'_i$ back into $\mathcal{M}$ to assist similar future tasks, Experience extracts a separate rule set $\mathcal{K}$ that serves as a policy prior for unseen scenarios, marking a fundamental shift from trajectory-local refinement to cross-trajectory abstraction.}
\label{tab:reflection_vs_experience}
\end{table*}

\section{Transformative Experience}\label{Transformative Experience}
Following the exposition of the evolutionary trajectory for memory mechanisms, we address \textbf{RQ3: What changes does Experience bring?} In this section, we elucidate the distinct technical characteristics of experience as a novel stage in the development of memory mechanisms.

\subsection{Active Exploration}
Active exploration leverages memory mechanisms to transform LLM agents from passive recorders of information into collectors of experience driven by goals. In the stage of Experience, the core capability of memory mechanisms is no longer confined to the storage of history, but extends to the acquisition of valuable experience through the active exploration of the environment. Here, exploration is framed as a memory-centric process, where prior experience guides its direction and its outcomes are abstracted back into memory.

\noindent\textbf{Exploration Mechanisms.}
The driving mechanisms for active exploration have transitioned from traditional strategies of random exploration toward more profound drivers of intrinsic motivation and feedback. Drivers based on signals of reward guide LLM agents to explore state spaces of greater value through the design and optimization of immediate reward functions~\citep{Zheng2024OnlineIR, Pan2025WonderWW, Sun2025CuriosityDrivenRL}; drivers based on curricula facilitate exploration tasks of increasing difficulty through the dynamic generation and adjustment of sequences for tasks~\citep{Wei2025WebAgentR1TW, Ahn2022DoAI}; and drivers based on reuse enable exploration of high efficiency through the abstraction and reuse of trajectories from history~\citep{Wang2025RAGENUS, Cai2025FLEXCA}.

\noindent\textbf{Exploration Dimensions.}
The core of active exploration resides in the utilization of memory mechanisms to facilitate the expansion of the boundaries of capability for LLM agents. This process can be categorized into three critical dimensions: exploration of breadth aims to alleviate cognitive deficiencies of LLM agents in unfamiliar environments, transforming memory into experience that is structured through mechanisms of curiosity~\citep{qi2025webrltrainingllmweb, Zhai2025AgentEvolverTE, Cheng2025WebATLASAL}; exploration of depth focuses on the extraction of skills of a high order within vertical tasks, driving the evolution of memory from the basic following of instructions to complex experiential strategies~\citep{Xia2025Agent0US, Liu2025Agent0VLES}; and exploration of strategy centers on the dynamic optimization of paths for decision making, leveraging the accumulation of experience to enhance the precision of decisions for LLM agents during planning over long horizons~\cite{Shi2025MonteCP, BIDOCHKO2026102740}.

\subsection{Cross-Trajectory Abstraction}
Cross-trajectory abstraction compresses isolated trajectories into universal patterns, transforming scattered and episodic experiences into stable priors for policy. This enables LLM agents to transcend specific sequences of actions and engage in decision making at higher dimensions of abstraction, which provides prospective guidance for tasks that are unknown and facilitates an understanding of underlying regularities.

\begin{figure}[t!] 
    \centering
    \includegraphics[width=1\linewidth]{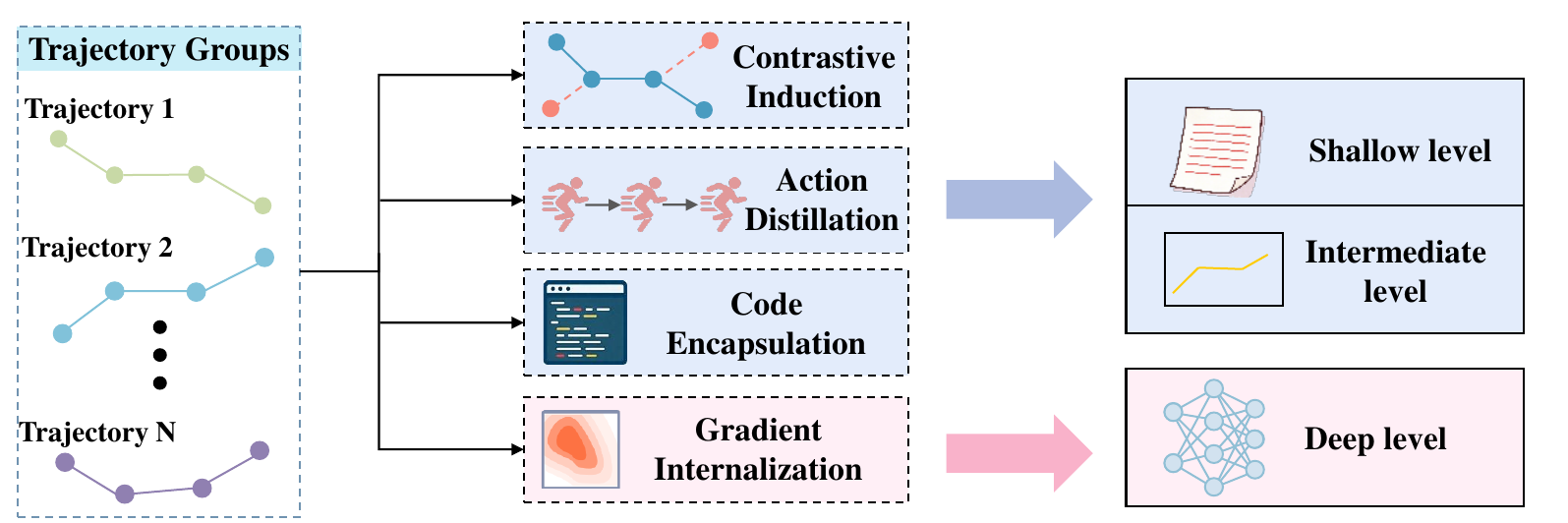}
    \caption{Overview of Cross-Trajectory Abstraction.}
    \label{fig:3}
    \vspace{-0.3cm}
\end{figure}

\noindent\textbf{Abstraction Mechanisms.}
The mechanism for abstraction serves as the core operator for the transformation of groups of raw interaction trajectories into experience that is universal. In contrast to mechanisms for reflection that focus on the correction of errors within single trajectories, abstraction during the stage of experience emphasizes the execution of inductive operations across trajectories. According to its operational logic, this process includes contrastive induction, which utilizes the opposition between successful and failed trajectories to delineate the boundaries of policy with precision~\citep{Forouzandeh2025LearningHP, He2024IDEAET}; the distillation of actions of fine granularity into patterns of thought of a high order through the chunking and aggregation of behavioral sequences across multiple levels of granularity~\citep{Fang2025MempEA, Latimer2025HindsightI2}; the encapsulation of recurring patterns of behavior into program functions that are reusable by leveraging the compositionality of code~\citep{Yang2025AutomatedSD, Zhang2025MemEvolveMO, Wang2025InducingPS}; and the internalization of groups of trajectories into the parameters of the model through techniques for fine-tuning~\citep{Ding2025RethinkingET, Chen2025InternalizingWM}.

\noindent\textbf{Abstraction Granularity.}
The hierarchy of abstraction for memory mechanisms determines the boundaries for generalization and the degree of interpretability for experience. Based on the degree to which the results of abstraction deviate from original trajectories, these can be categorized into three progressive levels: abstraction at the shallow level retains a portion of semantic logic, utilizing “rules” described in natural language as experience~\citep{Cao2025RememberMR, Chen2025SWEExpES, Wei2025EvoMemoryBL, Hayashi2025SelfAbstractionFG}; abstraction at the intermediate level completely removes redundancies of natural language, extracting only modular skeletons for execution as experience~\citep{Wang2024AgentWM, Liu2025ContextualER, Yu2025PolySkillLG}; and abstraction at the deep level compresses the distribution of trajectories into the weights of the model, enabling the complete transformation of experience into intuition for decision making~\citep{Cheng2025AgentR1TP, Luo2025AgentLT, Wang2025ReinforcementLF}.

{\setlength{\abovedisplayskip}{0.1cm}
\setlength{\belowdisplayskip}{0.1cm}
\begin{summarybox}{Summary \& Ideas - Transformative Experience}
    \begin{itemize}[leftmargin=1.8em, nosep] 
        \item[{\makebox[1em][c]{\textcolor{Summary}{\faBook}}}] Within the novel stage of memory mechanisms characterized by Experience, active exploration and abstraction across trajectories represent the two primary features of this phase.
        \item[{\makebox[1em]{\textcolor{Ideas}{\faLightbulb}}}] The feedback loop between exploration and abstraction serves as the central engine that drives the autonomous and continuous evolution of LLM agents during the phase of experience.
        \item[{\makebox[1em]{\textcolor{Ideas}{\faLightbulb}}}] Memories characterized by a high degree of abstraction during the phase of experience also enable agents to circumvent challenges frequently associated with conventional memory mechanisms~\citep{Xiong2025HowMM,Gharat2025FromPT, srivastava2025memorygraftpersistentcompromisellm}.
    \end{itemize}
\end{summarybox}}

\section{Future Directions}\label{Future Direction}
In this section, we discuss emerging prospects and promising directions for memory mechanisms for LLM agents.

\noindent\textbf{Active Memory Perception.}
Currently, certain memory mechanisms still utilize modes of passive triggering, which necessitates that LLM agents perform indiscriminate retrieval of a significant portion of the memory repository~\citep{Wang2025OMemOM, Rasmussen2025ZepAT}. More importantly, the persistent retrieval of irrelevant or obsolete memories can disrupt the coherence in reasoning of the LLM agent~\cite{Xu2025AMEMAM,Tan2025InPA}. Recent work has begun to address this challenge through autonomous retrieval controllers~\citep{Du2025MemR3MR}. Future research demands that memory mechanisms autonomously evaluate whether a task requires the introduction of additional memory and determine the specific type of memory to be integrated, ensuring that memory mechanisms function as resources that are invoked on demand.

\noindent\textbf{Organization of Working Memory.}
As the complexity of tasks and the horizons encountered by LLM agents continue to expand, the construction of working memory within tasks has emerged as a primary bottleneck. LLM agents must reconstruct trajectories from the past into memory intervals that are dynamic and plastic to facilitate the effective allocation of attention~\citep{Hu2024HiAgentHW, Luo2025BeyondCL}. Future research may focus on the isolation of interval memory, the retrospective integration of critical nodes for decision making, and the adaptive pruning of working memory~\citep{Sun2025ScalingLL, zhang2025memoryactionautonomouscontext, Nan2025NemoriSA}.

\noindent\textbf{Benchmark for Experience.}
Existing datasets primarily evaluate the capacity for retrieval and denoising of memory within the stages of storage and reflection, whereas the evaluation of the capacity for abstraction and generalization during the stage of experience remains significantly insufficient (Appendix~\S\ref{Datasets}). The assessment of the lifecycle of experience is closely linked to the capacity for meta-learning in LLM agents, which is essential for the realization of systems for self evolution based on active generalization~\citep{Behrouz2024TitansLT, Wei2025EvoMemoryBL}. Consequently, the path of evolution we propose provides a valuable foundation for guiding the development of these benchmarks.

\noindent\textbf{Distributed Shared Memory.}
Collaboration among multiple agents represents the essential pathway toward the realization of “Organizations.” This establishes distributed shared memory as central to current research~\cite{doi:10.36227/techrxiv.176539617.79044553/v1}. At present, mechanisms for shared memory rely primarily on communication through explicit dialogue, which is not only constrained by bottlenecks in bandwidth but is also prone to the introduction of noise during the process of exchange~\citep{Tran2025MultiAgentCM, Liao2025AgentMasterAM, Zou2025LatentCI}. To overcome current communication constraints, future efforts should prioritize the development of consensus memory systems. These systems aim to achieve efficient synchronization between individual perspectives and collective knowledge, thereby fostering a more agile process of socialized experience evolution~\citep{Yuen2025IntrinsicMA, Rezazadeh2025CollaborativeMM}.

\noindent\textbf{Multimodal Memory.}
Multimodal memory represents a significant direction for the future development of memory mechanisms for LLM agents. This direction requires the integration of states of visual perception, processes of linguistic reasoning, and other perceptual modalities into memory units characterized by unified temporality and semantics~\citep{Liu2025MemVerseMM, Zhou2025M2PAAM, He2025RSAgentLT}. For embodied intelligence in particular, the integrity of internal models of the world directly influences the planning and execution of tasks~\citep{Feng2025EmbodiedAF, Long2025ASL}. To achieve this objective, existing research explores novel memory mechanisms by investigating multimodal abstraction, temporal alignment across modalities, and the efficient consolidation of memory. A detailed exposition of current approaches and open challenges is provided in 
Appendix~\S\ref{Extended Discussion on Multimodal Memory Mechanisms}.

\section{Conclusion}
This survey provides a systematic review of memory mechanisms for LLM agents, establishing an evolutionary framework that encompasses three progressive stages: storage, reflection, and experience. Our analysis demonstrates that memory evolution is not merely the expansion of storage capacity but fundamentally involves the enhancement of information density and a transformation across cognitive abstraction dimensions. By introducing mechanisms such as active exploration and cross-trajectory abstraction, memory mechanisms within the experience stage enable agents to transcend situational constraints and acquire transferable behavioral experience. We hope this survey assists the community in designing more advanced memory mechanisms, guiding LLM agents toward the realization of true general artificial intelligence.

\section*{Limitations}
This survey provides a comprehensive qualitative analysis of memory mechanisms for LLM agents; however, we acknowledge several limitations that warrant discussion.

\noindent\textbf{Lack of Direct Quantitative Comparison.}
This survey adopts a qualitative analytical framework and lacks a comprehensive performance comparison of memory mechanisms. This is because the design objectives differ across the three stages of storage, reflection, and experience, and no unified benchmark currently exists for comprehensive evaluation across all stages. Moreover, variations in foundation models, environments, and prompts across original studies render direct numerical comparison potentially misleading.

\noindent\textbf{Relation to Established Learning Paradigms.}
The experience stage, particularly implicit experience, intersects with fine-tuning, reinforcement learning, and meta-learning at a technical level. This taxonomy does not position experience as an entirely novel learning paradigm; rather, it emphasizes how these established techniques are deployed within memory-centric LLM agent architectures, serving as a critical intermediary between interaction trajectories and parameter updates.

\noindent\textbf{Temporal Coverage and Recency Bias.}
Research on memory mechanisms for LLM agents has experienced rapid growth from 2024 to 2025, with the experience stage emerging as a coherent research direction only in the latter half of 2025. This temporal distribution is reflected in the coverage of this survey and brings two methodological implications: (i) early influential works may not have received attention commensurate with their historical contributions, and (ii) some recent preprints included in this survey have not yet undergone formal peer review. To balance academic rigor with timeliness, this survey prioritizes works that propose novel architectures or demonstrate reproducible results.

\section*{Acknowledgment}
This work is partially supported by National Natural Science Foundation of China Young Scientists Fund (No. 62206233) and RMGS (2025 First Processing Cycle).

\bibliography{anthology}

\appendix
\renewcommand{\theequation}{\thesection.\arabic{equation}}
\setcounter{equation}{0}
\section{Overview}\label{Overview}
We first provide a formal definition for the operational framework of LLM agents and the evolutionary paradigms of memory mechanisms. Furthermore, we categorize the primary drivers for this evolution into three dimensions, emphasizing that these driving forces constitute the fundamental factors supporting transformations in memory mechanisms and the inherent capabilities of LLM agents. Based on the depth to which historical trajectories are utilized, this survey proposes an evolutionary framework consisting of three distinct stages:
\begin{itemize}[leftmargin=*, leftmargin=1.5em, rightmargin=1em]
    \item \textbf{Storage:} As the foundational layer of evolution, this stage focuses on the faithful preservation of trajectories from interactions over a long duration to address constraints regarding the memory capacity of LLM agents.
    \item \textbf{Reflection:} Through the introduction of loops for dynamic evaluation, the memory mechanism transitions from a recorder of information to an evaluator, thereby mitigating issues related to hallucinations and logical errors within the memory of LLM agents.
    \item \textbf{Experience:} Representing the highest level of cognition, this stage employs abstraction across multiple trajectories to extract behavioral patterns of a higher order. This process compresses redundant memory into heuristic strategies that are transferable and reusable.
\end{itemize}
Furthermore, we provide an in-depth discussion on two pivotal technological shifts required for memory mechanisms to advance toward the stage of experience: active exploration and abstraction across trajectories. These advancements enable LLM agents to transition from passive recipients of information to collectors of experience driven by specific goals, thereby enhancing the capacity for proactive generalization in tasks that are unknown. Finally, this survey discusses several valuable directions for the future of memory mechanisms.

\noindent\textbf{Summary of Contribution.}
As a survey, our primary objective is the synthesis and analysis of existing research, while providing novel insights and perspectives for researchers who aim to understand and design memory mechanisms. We believe that our work offers significant novelty in the following aspects:





\begin{itemize}[leftmargin=*, leftmargin=1.5em, rightmargin=1em]
    \item \textbf{Scope \& Coverage:} To address the absence of a perspective on evolution and the significant fragmentation in contemporary research on memory mechanisms, this survey provides a comprehensive overview that is forward-looking. This work encompasses research that has been overlooked, the most recent advancements, and theoretical perspectives of a broader nature.
    \item \textbf{Organization \& Structure:} This survey constructs an evolutionary framework in three stages to organize the manuscript. On this basis, we systematically delineate the drivers and pathways for the development of memory mechanisms, as well as characteristics at the frontier. This perspective provides novel insights for research within this domain.
    \item \textbf{Insights \& Critical Analysis:} This survey provides original interpretations and an in-depth analysis of the existing literature. For instance, we propose a taxonomy from an evolutionary perspective, using the degree of utilization for trajectories of past interaction as a benchmark. Furthermore, we summarize two pivotal characteristics of memory mechanisms in the stage of experience and identify several issues that remain underexplored or unresolved.
    \item \textbf{Timeliness \& Relevance:} In the inaugural year of LLM agents, this work represents the first survey to systematically examine memory mechanisms from a perspective of evolution, capturing research at the frontier through 2025. It addresses the urgent necessity for adaptation and learning as agents encounter the real world for the first time. Through the synthesis of existing literature, we provide a new foundation for further exploration and innovation in this critical field.
\end{itemize}

Overall, this survey offers a novel perspective by providing comprehensive coverage and an innovative classification of memory evolution. We anticipate that these contributions will bridge the knowledge gap regarding LLM agent memory mechanisms and offer a foundational resource for future research.

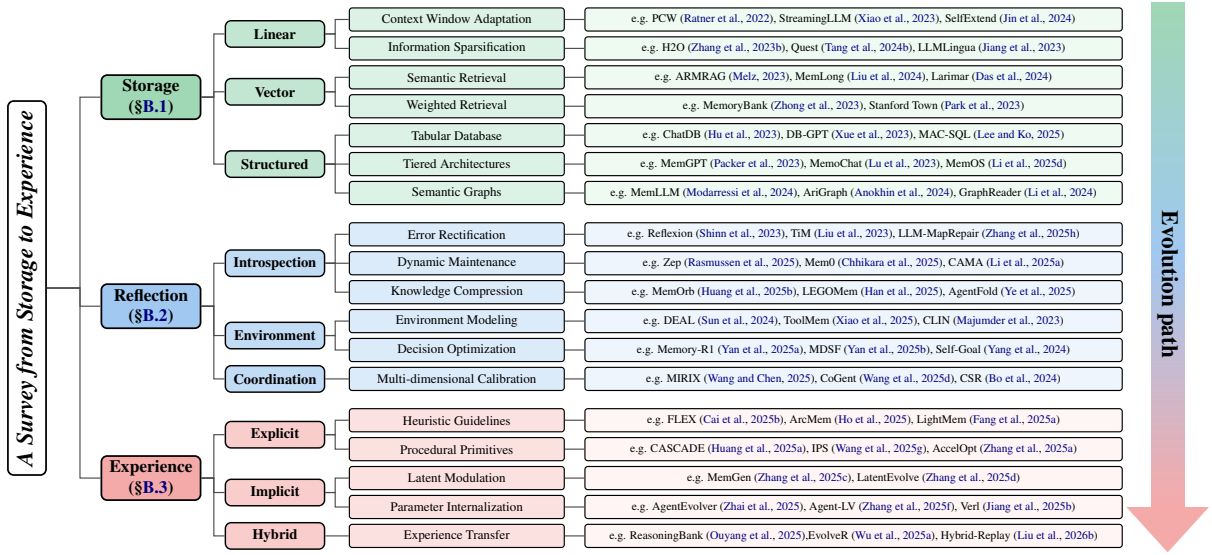
\begin{figure*}[!h]
\centering
\resizebox{1.0\textwidth}{!}{%
\begin{tikzpicture}[
    level1/.style={draw, line width=0.6pt, rounded corners=2pt, minimum width=1.2cm, minimum height=0.45cm, fill=memoryorange, font=\fontsize{6}{7}\selectfont\bfseries, align=center, inner sep=1.5pt},
    level2/.style={draw, line width=0.5pt, rounded corners=2pt, minimum width=1.2cm, minimum height=0.32cm, fill=memoryorangelight, font=\fontsize{5}{6}\selectfont\bfseries, align=center, inner sep=1pt},
    level3/.style={draw, line width=0.4pt, rounded corners=1pt, minimum width=2.6cm, minimum height=0.28cm, fill=memoryorangelighter, font=\fontsize{4.5}{5.5}\selectfont, align=center, inner sep=1pt},
    level4/.style={draw, line width=0.4pt, rounded corners=1pt, minimum width=6.5cm, minimum height=0.28cm, fill=memoryorangelightest, font=\fontsize{4}{5.5}\selectfont, align=center, inner sep=1.5pt},
    line/.style={line width=0.4pt, draw=black!70}, 
]

\def\vstep{0.35} 

\def\colTitle{-1.6}  
\def\colOne{-0.1}
\def\colTwo{1.4}
\def\colThree{3.6}
\def\colFour{8.4}

\def\gapL{0.12}
\def\gapM{0.15}

\def\arrowX{12}

\node[level3] (cwa) at (\colThree, 2.80) {Context Window Adaptation};
\node[level4] (cwa-ex) at (\colFour, 2.80) {e.g. PCW~\cite{Ratner2022ParallelCW}, StreamingLLM~\cite{Xiao2023EfficientSL}, SelfExtend~\cite{Jin2024LLMML}};

\node[level3] (ics) at (\colThree, 2.45) {Information Sparsification};
\node[level4] (ics-ex) at (\colFour, 2.45) {e.g. H2O~\cite{Zhang2023H2OHO}, Quest~\cite{Tang2024QuestQS}, LLMLingua~\cite{Jiang2023LLMLinguaCP}};

\node[level2] (linear) at (\colTwo, 2.625) {Linear};

\node[level3] (sr) at (\colThree, 2.10) {Semantic Retrieval};
\node[level4] (sr-ex) at (\colFour, 2.10) {e.g. ARMRAG~\cite{Melz2023EnhancingLI}, MemLong~\cite{Liu2024MemLongMR}, Larimar~\cite{Das2024LarimarLL}};

\node[level3] (wr) at (\colThree, 1.75) {Weighted Retrieval};
\node[level4] (wr-ex) at (\colFour, 1.75) {e.g. MemoryBank~\cite{Zhong2023MemoryBankEL}, Stanford Town~\cite{Park2023GenerativeAI}};

\node[level2] (vector) at (\colTwo, 1.925) {Vector};

\node[level3] (tdb) at (\colThree, 1.40) {Tabular Database};
\node[level4] (tdb-ex) at (\colFour, 1.40) {e.g. ChatDB~\cite{Hu2023ChatDBAL}, DB-GPT~\cite{Xue2023DBGPTED}, MAC-SQL~\cite{Lee2025TrainingAT}};

\node[level3] (ta) at (\colThree, 1.05) {Tiered Architectures};
\node[level4] (ta-ex) at (\colFour, 1.05) {e.g. MemGPT~\cite{Packer2023MemGPTTL}, MemoChat~\cite{Lu2023MemoChatTL},  MemOS~\cite{Li2025MemOSAM}};

\node[level3] (sg) at (\colThree, 0.70) {Semantic Graphs};
\node[level4] (sg-ex) at (\colFour, 0.70) {e.g. MemLLM~\cite{Modarressi2024MemLLMFL}, AriGraph~\cite{Anokhin2024AriGraphLK}, GraphReader~\cite{Li2024GraphReaderBG}};

\node[level2] (structured) at (\colTwo, 1.05) {Structured};

\node[level1] (memory) at (\colOne, 1.86) {Storage \\ (\S\ref{Storage})};    

\node[level3, fill=reflectiveyellowlighter] (er) at (\colThree, 0.20) {Error Rectification};
\node[level4, fill=reflectiveyellowlightest] (er-ex) at (\colFour, 0.20) {e.g. Reflexion~\cite{Shinn2023ReflexionLA}, TiM~\cite{Liu2023ThinkinMemoryRA}, LLM-MapRepair~\cite{Zhang2025ConstructingCS}};

\node[level3, fill=reflectiveyellowlighter] (dm) at (\colThree, -0.15) {Dynamic Maintenance};
\node[level4, fill=reflectiveyellowlightest] (dm-ex) at (\colFour, -0.15) {e.g. Zep~\cite{Rasmussen2025ZepAT}, Mem0~\cite{Chhikara2025Mem0BP}, CAMA~\cite{Li2025CAMAC}};

\node[level3, fill=reflectiveyellowlighter] (kc) at (\colThree, -0.50) {Knowledge Compression};
\node[level4, fill=reflectiveyellowlightest] (kc-ex) at (\colFour, -0.50) {e.g. MemOrb~\cite{Huang2025MemOrbAP}, LEGOMem~\cite{Han2025LEGOMemMP}, AgentFold~\cite{Ye2025AgentFoldLW}};

\node[level2, fill=reflectiveyellowlight] (introspective) at (\colTwo, -0.15) {Introspection};

\node[level3, fill=reflectiveyellowlighter] (em) at (\colThree, -0.85) {Environment Modeling};
\node[level4, fill=reflectiveyellowlightest] (em-ex) at (\colFour, -0.85) {e.g. DEAL~\cite{Sun2024EnhancingAL}, ToolMem~\cite{Xiao2025ToolMemEM}, CLIN~\cite{Majumder2023CLINAC}};

\node[level3, fill=reflectiveyellowlighter] (do) at (\colThree, -1.20) {Decision Optimization};
\node[level4, fill=reflectiveyellowlightest] (do-ex) at (\colFour, -1.20) {e.g. Memory-R1~\cite{Yan2025MemoryR1EL}, MDSF~\cite{Yan2025MemoryDrivenSF}, Self-Goal~\cite{Yang2024SelfGoalYL}};

\node[level2, fill=reflectiveyellowlight] (environmental) at (\colTwo, -1.025) {Environment};

\node[level3, fill=reflectiveyellowlighter] (mdc) at (\colThree, -1.55) {Multi-dimensional Calibration};
\node[level4, fill=reflectiveyellowlightest] (mdc-ex) at (\colFour, -1.55) {e.g. MIRIX~\cite{Wang2025MIRIXMM}, CoGent~\cite{Wang2025GraphCogentML}, CSR~\cite{Bo2024ReflectiveMC}};

\node[level2, fill=reflectiveyellowlight] (Coordination) at (\colTwo, -1.55) {Coordination};

\node[level1, fill=reflectiveyellow] (reflective) at (\colOne, -0.67) {Reflection \\ (\S\ref{Reflection})};

\node[level3, fill=experiencegreenlighter] (hg) at (\colThree, -2.05) {Heuristic Guidelines};
\node[level4, fill=experiencegreenlightest] (hg-ex) at (\colFour, -2.05) {e.g. FLEX~\cite{Cai2025FLEXCA}, ArcMem~\cite{Ho2025ArcMemoAR}, LightMem~\cite{Fang2025LightMemLA}};

\node[level3, fill=experiencegreenlighter] (pp) at (\colThree, -2.40) {Procedural Primitives};
\node[level4, fill=experiencegreenlightest] (pp-ex) at (\colFour, -2.40) {e.g. CASCADE~\cite{Huang2025CASCADECA}, IPS~\cite{Wang2025InducingPS}, AccelOpt~\cite{Zhang2025AccelOptAS}};

\node[level2, fill=experiencegreenlight] (explicit) at (\colTwo, -2.225) {Explicit};

\node[level3, fill=experiencegreenlighter] (ew) at (\colThree, -2.75) {Latent Modulation};
\node[level4, fill=experiencegreenlightest] (ew-ex) at (\colFour, -2.75) {e.g. MemGen~\cite{Zhang2025MemGenWG}, LatentEvolve~\cite{Zhang2025LatentEvolveST}};

\node[level3, fill=experiencegreenlighter] (pi) at (\colThree, -3.10) {Parameter Internalization};
\node[level4, fill=experiencegreenlightest] (pi-ex) at (\colFour, -3.10) {e.g. AgentEvolver~\cite{Zhai2025AgentEvolverTE}, Agent-LV~\cite{Zhang2025AgentLV}, Verl~\cite{Jiang2025VerlToolTH}};

\node[level2, fill=experiencegreenlight] (implicit) at (\colTwo, -2.925) {Implicit};

\node[level3, fill=experiencegreenlighter] (et) at (\colThree, -3.45) {Experience Transfer};
\node[level4, fill=experiencegreenlightest] (et-ex) at (\colFour, -3.45) {e.g. ReasoningBank~\cite{Ouyang2025ReasoningBankSA},EvolveR~\cite{Wu2025EvolveRSL}, Hybrid-Replay~\cite{liu2026exploratorymemoryaugmentedllmagent}};

\node[level2, fill=experiencegreenlight] (hybrid) at (\colTwo, -3.45) {Hybrid};

\node[level1, fill=experiencegreen] (experience) at (\colOne, -2.75) {Experience \\ (\S\ref{Experience})};  

\node[draw, line width=0.6pt, black, rounded corners=2pt, inner sep=3pt, font=\fontsize{8}{10}\selectfont\bfseries\itshape, rotate=90, anchor=center] (titlenode) at (\colTitle, -0.45) {A Survey from Storage to Experience};

\draw[line] (titlenode.south) -- ++(0.4,0) |- (memory.west);
\draw[line] (titlenode.south) -- ++(0.4,0) |- (reflective.west);
\draw[line] (titlenode.south) -- ++(0.4,0) |- (experience.west);

\draw[line] (memory.east) -- ++(\gapM,0) |- (linear.west);
\draw[line] (memory.east) -- ++(\gapM,0) |- (vector.west);
\draw[line] (memory.east) -- ++(\gapM,0) |- (structured.west);

\foreach \child in {cwa, ics} \draw[line] (linear.east) -- ++(\gapL,0) |- (\child.west);
\foreach \child in {sr, wr} \draw[line] (vector.east) -- ++(\gapL,0) |- (\child.west);
\foreach \child in {tdb, ta, sg} \draw[line] (structured.east) -- ++(\gapL,0) |- (\child.west);

\draw[line] (reflective.east) -- ++(\gapM,0) |- (introspective.west);
\draw[line] (reflective.east) -- ++(\gapM,0) |- (environmental.west);
\draw[line] (reflective.east) -- ++(\gapM,0) |- (Coordination.west);

\foreach \child in {er, dm, kc} \draw[line] (introspective.east) -- ++(\gapL,0) |- (\child.west);
\foreach \child in {em, do} \draw[line] (environmental.east) -- ++(\gapL,0) |- (\child.west);
\draw[line] (Coordination.east) -- ++(\gapL,0) -- (mdc.west);

\draw[line] (experience.east) -- ++(\gapM,0) |- (explicit.west);
\draw[line] (experience.east) -- ++(\gapM,0) |- (implicit.west);
\draw[line] (experience.east) -- ++(\gapM,0) |- (hybrid.west);

\foreach \child in {hg, pp} \draw[line] (explicit.east) -- ++(\gapL,0) |- (\child.west);
\foreach \child in {ew, pi} \draw[line] (implicit.east) -- ++(\gapL,0) |- (\child.west);
\draw[line] (hybrid.east) -- ++(\gapL,0) -- (et.west);

\foreach \start/\end in {cwa/cwa-ex, ics/ics-ex, sr/sr-ex, wr/wr-ex, tdb/tdb-ex, ta/ta-ex, sg/sg-ex, 
                         er/er-ex, dm/dm-ex, kc/kc-ex, em/em-ex, do/do-ex, mdc/mdc-ex, 
                         hg/hg-ex, pp/pp-ex, ew/ew-ex, pi/pi-ex, et/et-ex} {
    \draw[line] (\start.east) -- (\end.west);
}

\begin{scope}[on background layer]
    \shade[top color=arrowtop, bottom color=arrowmiddle]
        (\arrowX, 3.0) rectangle ++(0.4, -3.05);
    \shade[top color=arrowmiddle, bottom color=arrowbottom]
        (\arrowX, -0.05) rectangle ++(0.4, -3.05);
    \fill[arrowbottom]
        (\arrowX-0.3, -3.1) -- (\arrowX+0.2, -3.65) -- (\arrowX+0.7, -3.1) -- cycle;
\end{scope}
\node[rotate=-90, font=\fontsize{8}{9}\selectfont\bfseries, anchor=center, text=black] 
    at (\arrowX+0.2, -0.4) {Evolution path};

\end{tikzpicture}
}%
\caption{Taxonomy of the LLM agent memory mechanisms.}
\label{fig:taxonomy}
\end{figure*}

\section{Detail within the Evolutionary Path}\label{Detail within the Evolutionary Path}
Due to space constraints in the main text, we provide detailed exposition of representative works within each stage of the memory mechanism evolution in this section. For each work, we describe its core contribution, technical mechanism, and position within the evolutionary trajectory.

\subsection{Storage}\label{Storage}
The primary objective of the storage phase is the precise preservation of trajectories to the maximum extent possible, which enables LLM agents to maintain an accurate perception of both internal and external states~\cite{Xi2023TheRA}. Although the memory mechanisms of the storage phase provide the context necessary for continuity and reasoning, they remain inherently susceptible to contamination from the stochasticity and hallucinations of the underlying model. Prior research has addressed the requirements for the writing, management, and retrieval of memory within various environments by constructing memory architectures across three technical categories: linear, vector, and structured.

\noindent\textbf{Linear.}\label{Linear}
Linear storage represents the most primitive and intuitive form of memory mechanisms. It treats the records of interaction for LLM agents as a continuous stream of tokens arranged in chronological order, managing memory within the context window through strict adherence to a strategy of First-In, First-Out (FIFO). The work in this phase can be categorized into two components: adjustment of the context window and sparsification of information.

\begin{itemize}[leftmargin=*, leftmargin=1.5em, rightmargin=1em]
    \item \textbf{Context Window Adaptation:} Context window adaptation techniques seek to extend the usable input length of LLMs by modifying attention mechanisms, positional encoding schemes, or input structures. Representative approaches include optimizing intrinsic attention computation ~\cite{Xiao2023EfficientSL}, remapping positional encodings to enable longer sequences ~\cite{Jin2024LLMML}, and restructuring inputs to mitigate length constraints ~\cite{Ratner2022ParallelCW}. These methods expand raw storage capacity but do not alter the semantics of stored trajectories.
    \item \textbf{Information Sparsification:} Information sparsification treats memory compression as a mechanical denoising process independent of agent reflection. Methods typically rely on statistical or attention-based heuristics to remove low-utility tokens. For example,~\citet{Zhang2023H2OHO} evicts tokens based on cumulative attention scores, while~\citet{Tang2024QuestQS} and~\citet{Xiao2024InfLLMTL} retrieve salient memory blocks via query–key similarity.~\citet{Jiang2023LLMLinguaCP} further identifies redundant segments through perplexity estimation. While effective for efficiency, these methods operate without semantic abstraction.
\end{itemize}

\noindent\textbf{Vector.}\label{Vector}
Vector storage mitigates the constraints of capacity for memory storage by encoding interaction trajectories into spaces of high dimensionality. However, it also introduces a novel challenge: the efficient retrieval of memories relevant to the task from massive repositories. Consequently, the focus of research has transitioned toward the optimization of retrieval. We categorize these methodologies into two classes: semantic retrieval and weighted retrieval.

\begin{itemize}[leftmargin=*, leftmargin=1.5em, rightmargin=1em]
    \item \textbf{Semantic Retrieval:} Semantic retrieval constitutes the foundational approach to vector memory, where relevance is determined by geometric proximity in embedding space. ~\citet{Melz2023EnhancingLI} retrieves historical reasoning chains via semantic alignment, while~\citet{Liu2024MemLongMR} integrates fine-grained retrieval-attention during decoding to sustain long-context reasoning.~\citet{Das2024LarimarLL} further internalizes episodic memory into a latent matrix, enabling one-shot read–write operations. Despite improved recall, these methods treat retrieved content as flat historical evidence.
    \item \textbf{Weighted Retrieval:} Weighted retrieval extends semantic similarity by assigning differentiated importance to memories using multi-dimensional scoring signals.~\citet{Zhong2023MemoryBankEL} models temporal decay via the Ebbinghaus Forgetting Curve, while ~\citet{Park2023GenerativeAI} retrieves memories based on a weighted combination of relevance, recency, and importance. Such mechanisms improve prioritization but remain retrieval-centric rather than abstraction-driven.
\end{itemize}

\noindent\textbf{Structured.}\label{Structured}
Structured storage preserves memory through predefined structures of relationships. This paradigm emphasizes the integrity and enforcement of knowledge within memory, which facilitates precise operations, complex reasoning based on logic, and efficient retrieval across multiple hops. Based on the method of organization, we categorize these systems into three classes: tabular databases, tiered architectures and semantic graphs.

\begin{itemize}[leftmargin=*, leftmargin=1.5em, rightmargin=1em]
    \item \textbf{Tabular Database:} Database-backed memory systems leverage mature relational databases to store agent knowledge in structured tabular form. Early work frames databases as symbolic memory ~\cite{Hu2023ChatDBAL}, while subsequent approaches translate natural language queries into SQL via specialized controllers for secure and efficient access ~\cite{Xue2023DBGPTED}. Multi-agent extensions further distribute database construction and maintenance across specialized roles ~\cite{Lee2025TrainingAT}.
    \item \textbf{Tiered Architectures:} Tiered memory architectures draw inspiration from computer storage hierarchies and human cognition to balance capacity and access latency. MemGPT ~\cite{Packer2023MemGPTTL} introduces a dual-layer design separating main and external context, enabling virtual context expansion. Cognitive-inspired systems such as SWIFT–SAGE ~\cite{Lin2023SwiftSageAG} dynamically adjust retrieval intensity, while streaming-update architectures maintain long-term stability without exhaustive retrieval ~\cite{Zhou2023RecurrentGPTIG,Lu2023MemoChatTL}. 
    \item \textbf{Semantic Graphs:} Graph memory represents interaction histories as networks of entities and relations, enabling structured reasoning beyond flat storage. Triplet-based extraction supports precise updates and retrieval ~\cite{Modarressi2024MemLLMFL}, while neuro-symbolic approaches integrate logical constraints into graph representations ~\cite{Wang2024SymbolicWM}. Graph-based world models further support environment-centric reasoning ~\cite{Anokhin2024AriGraphLK}, and coarse-to-fine traversal over text graphs enables efficient long-context retrieval ~\cite{Zhou2023RecurrentGPTIG,Lu2023MemoChatTL}. 
\end{itemize}

\subsection{Reflection}\label{Reflection}
Although the storage stage explores diverse methods for the preservation of memory to ensure the consistency of LLM agents over the long term, these approaches do not fundamentally address the quality of memory. Raw trajectories of interaction inevitably conflate successful sequences with hallucinations, errors in logic, and attempts that are invalid~\cite{Zhang2023HowLM, Ghasemabadi2025CanLP, Zhang2025ReplayFA}. Without the application of evaluation, the passive storage of all trajectories leads to an accumulation of errors and the repetition of failures. The reflection stage incorporates introspection, the environment, and coordination as signals for feedback to rectify and denoise historical trajectories, thereby producing memory of higher quality.

\noindent\textbf{Introspection.}\label{Introspection}
Introspective reflection represents an internal cognitive process that utilizes the LLM agent’s own knowledge to evaluate, refine, and restructure memory without the need for external feedback. Current research achieves introspection through three distinct functional pathways: error rectification, dynamic maintenance, and knowledge compression.

\begin{itemize}[leftmargin=*, leftmargin=1.5em, rightmargin=1em]
    \item \textbf{Error Rectification:} targets hallucinations and multi-step reasoning errors by verifying and repairing stored trajectories through self-critique. \citet{Shinn2023ReflexionLA} introduces Reflexion, which prompts agents to reflect on failed trajectories and distill corrective feedback into textual memory. This mechanism enables systematic error correction and sustained performance improvement across episodes, establishing introspective reflection as a central mechanism rather than a peripheral heuristic.Building on this paradigm,~\citet{Liu2023ThinkinMemoryRA} introduces a post-reasoning verification stage to retain only validated memories, while~\citet{Zhang2025ConstructingCS} detects contradictory or erroneous segments through introspective consistency checks, thereby limiting error accumulation and propagation.
    \item \textbf{Dynamic Maintenance:} Dynamic maintenance focuses on lifecycle management of memory content.~\citet{Li2025CAMAC} incrementally updates internal knowledge schemas via clustering, while~\citet{Rasmussen2025ZepAT} and ~\citet{Chhikara2025Mem0BP} maintain continuity by parsing and updating structured entity relations. At the system level, rule-based controllers inspired by operating systems strategically update and persist core memories ~\cite{Packer2023MemGPTTL,Zhou2025MEM1LT,Kang2025MemoryOO}.
    \item \textbf{Knowledge Compression:} Knowledge compression distills high-dimensional trajectories into compact and reusable representations.~\citet{Huang2025MemOrbAP} generates structured reflections to extract coherent character profiles, while~\citet{Han2025LEGOMemMP} decomposes interaction sequences into modular procedural memories. Multi-granularity abstraction further aligns distilled memories with task demands ~\cite{Tan2025InPA,Yang2025CoarsetoFineGM}, and context-folding techniques preserve working-context efficiency during reasoning ~\cite{Sun2025ScalingLL,Ye2025AgentFoldLW,Li2025DeepAgentAG}.
\end{itemize}

\noindent\textbf{Environment.}\label{Environment}
While introspective reflection leverages knowledge within the model to refine memory, it inherently carries a risk of inconsistency with factual reality. To mitigate this risk, reflection from the environment utilizes outcomes in the real world to actively optimize behavior and calibrate the internal knowledge of the model. Current research primarily proceeds along two trajectories: environment modeling and decision optimization.

\begin{itemize}[leftmargin=*, leftmargin=1.5em, rightmargin=1em]
    \item \textbf{Environment Modeling:} Environmental modeling aligns internal memory with dynamic external conditions such as environments, tools, and user preferences.~\citet{Sun2024EnhancingAL} enables agents to infer and validate world rules from demonstrations, while~\citet{Xiao2025ToolMemEM} summarizes tool behavior from execution outcomes. Preference-aware updates integrate short-term variation with long-term trends~\cite{Sun2025PreferenceAwareMU}, and EM-based formulations ensure memory consistency under distribution shifts~\cite{Yin2024ExplicitML}.
    \item \textbf{Decision Optimization:} Decision optimization treats memory management as a learnable policy guided by environmental rewards or execution feedback.~\citet{Yan2025MemoryR1EL} learns discrete actions from outcome-based rewards, while~\citet{Yan2025MemoryDrivenSF} refines memory quality using value-annotated decision trajectories. For complex planning, interaction feedback is used to validate and prune goal hierarchies~\cite{Yang2024SelfGoalYL}.
\end{itemize}

\noindent\textbf{Coordination.}\label{Coordination}
Collaborative reflection leverages the specialization of roles and mechanisms for consensus within systems of multiple agents to extend the process of reflection from the level of the individual to that of the collective. Through deliberation across multiple agents, this paradigm alleviates cognitive bottlenecks and hallucinations that are common in architectures with a single model during the processing of trajectories of complex memory.

\begin{itemize}[leftmargin=*, leftmargin=1.5em, rightmargin=1em]
    \item \textbf{Multi-dimensional Calibration:} Multi-dimensional calibration realizes distributed memory management through heterogeneous agent societies.~\citet{Wang2025MIRIXMM} coordinates core, episodic, and semantic memory modules to process multimodal long contexts.~\citet{Wang2025GraphCogentML} decomposes graph reasoning into perception, caching, and execution roles to reduce context loss. Narrative-level coherence is achieved by integrating episodic and semantic memories across agents ~\cite{Balestri2025NarrativeMI}. Moreover, ~\citet{Ozer2025MARMultiAgentRI} and~\citet{Bo2024ReflectiveMC} further enhance reasoning consistency and collaboration efficiency in agent societies by implementing collaborative reflection across diverse roles and personalized feedback mechanisms.
\end{itemize}

\subsection{Experience}\label{Experience}
Although reflection mechanisms mitigate hallucinations~\citep{Tian_Yan_Yang_Zhao_Chen_Wang_Luo_Ma_Song_2025} and noise through evaluation, their corrective efficacy remains at the level of trajectories and has not yet yielded knowledge at the level of strategy that is transferable~\citep{Shinn2023ReflexionLA, Renze2024SelfReflectionIL}. In addition, reflection focused on trajectories may lead to a linear expansion of the memory bank, which imposes a burden for inference and may potentially result in a characteristic of following trajectories~\citep{Hong2025EnhancingMR, Zhu2025WhereLA,Fu2025AgentRefineEA}. Consequently, memory mechanisms must transcend the limitations of reflecting on the past and move toward a stage of experience for the guidance of the future. At this stage, memory mechanisms abstract universal wisdom that is independent of context from clusters of trajectories; through this process, LLM agents~\citep{luo2026agentmathempoweringmathematicalreasoning,tian2025evolproveradvancingautomatedtheorem} truly liberate themselves from memory banks that are verbose and complex, achieving zero-shot transfer to scenarios that are unknown by means of skills or rules that are intuitive. Research on the stage of experience achieves prospective wisdom through the abstraction of experience in forms that are explicit, implicit, and hybrid.

\noindent\textbf{Explicit.}\label{Explicit}
Explicit experience abstracts patterns of knowledge that are readable by humans, editable, and generalizable from clusters of trajectories, framing experiential memory as insights of wisdom that allow for direct retrieval and reuse, analogous to the consultation of a reference manual or a library of functions. This methodology not only alleviates the pressure of inference but also provides LLM agents with capabilities for interpretability and self-evolution. Research on explicit experience can generally be categorized into Heuristic Guidelines and Procedural Primitives.

\begin{itemize}[leftmargin=*, leftmargin=1.5em, rightmargin=1em]
    \item \textbf{Heuristic Guidelines:} Heuristic guidelines serve to crystallize implicit intuition into explicit natural language strategies. In this domain, researchers focus on distilling experience into textual rules:~\citet{Ouyang2025ReasoningBankSA} abstracts key decision principles through contrastive analysis of successful and failed trajectories, while~\citet{Suzgun2025DynamicCT} proposes dynamically generated "prompt lists" for real-time heuristic guidance.~\citet{Xu2025SEDMSS} and~\citet{Hassell2025LearningFS} investigate rule induction from supervisory signals, achieving textual experience transfer via "cross-domain knowledge diffusion" and "semantic task guidance," respectively. To transcend linear text limitations in modeling complex dependencies, recent work shifts toward structured schemas.~\citet{Ho2025ArcMemoAR} and~\citet{Zhang2025GMemoryTH} abstract multi-turn reasoning traces into experience graphs, leveraging topological structures to capture logical dependencies and enable effective storage and reuse of collaboration patterns and high-level cognitive principles.
    Moreover,~\citet{Cai2025FLEXCA} organizes heuristic knowledge into modular and compositional units, enabling systematic reuse across tasks.
    \item \textbf{Procedural Primitives:} Procedural primitives represent the abstraction of complex reasoning chains into executable entities, designed to significantly reduce planning overhead.~\citet{Wang2025InducingPS} proposes a skill induction mechanism that encapsulates high-frequency action sequences into functions, enabling agents to invoke complex skills as atomic actions.~\citet{Zhang2025AccelOptAS} extends this executable paradigm to hardware optimization, enabling agents to accumulate kernel optimization skills that iteratively enhance accelerator performance. In this line of work,~\citet{Huang2025CASCADECA} enables the composition and cascading execution of such procedural primitives, allowing agents to construct complex behaviors through structured skill invocation.

\end{itemize}

\noindent\textbf{Implicit.}\label{Implicit}
Implicit experience eschews the retrieval of discrete text and abstracts the history of interactions into implicit priors, thereby addressing the overhead of inference and the constraints of context. Experiential memory is transformed into latent variables within spaces of high dimensionality or into parameters of the neural network. Based on the form of implementation for the transformation, implicit experience is categorized into two trajectories: Latent Modulation and Parameter Internalization.

\begin{itemize}[leftmargin=*, leftmargin=1.5em, rightmargin=1em]
    \item \textbf{Latent Modulation:} Latent modulation operates on the cognitive stream within continuous high-dimensional latent space. By encoding experience into latent variables or activation states, this paradigm "weaves" historical insights into current reasoning in a parameter-free manner, circumventing expensive parameter updates.~\citet{Zhang2025MemGenWG} introduces the MemGen framework, employing a "Memory Weaver" to dynamically generate and inject latent token sequences conditioned on current reasoning state.~\citet{Zhang2025LatentEvolveST} achieves smooth transfer from historical experience to current decision-making without altering static parameters, using alternating Fast Retrieval and Slow Integration within latent space.
    \item \textbf{Parameter Internalization:} Parameter Internalization transforms explicit trajectories into intrinsic capabilities within model weights. Through gradient updates, this mechanism instills adaptive priors into LLM agents, enabling effective navigation of complex environments. For context distillation,~\citet{Alakuijala2025MementoNM} proposes iterative distillation to internalize corrective hints into model weights.~\citet{Liu2025AnalyzingAI} converts business rules into model priors, alleviating retrieval overload in RAG systems, while~\citet{Zhai2025AgentEvolverTE} introduces "Experience Stripping," eliminating retrieval segments during training to force internalization of explicit experience into autonomous reasoning capabilities independent of external auxiliaries. For Reinforcement Learning (RL),~\citet{Zhang2025AgentLV} proposes a pioneering early experience paradigm, leveraging implicit world models and sub-reflective prediction to internalize trial-and-error experience into policy priors without extrinsic rewards.~\citet{Lyu2025FromCT} achieves strategic transformation from Reflection to Experience by applying RL to student-generated reflections.~\citet{Feng2025GroupinGroupPO} proposes group-based policy optimization for fine-grained experience internalization across multi-turn interactions.~\citet{Jiang2025VerlToolTH} establishes standardized alignment between RL and tool invocation, enhancing agents' capacity to transmute tool-use experience into intrinsic strategies.
\end{itemize}

\noindent\textbf{Hybrid.}\label{Hybrid}
Hybrid experience aims to transcend the dichotomy between explicit and implicit paradigms by establishing a dynamic "Accumulate-Internalize" cycle. This paradigm directly addresses the challenges of "Storage Explosion" and "Retrieval Latency" encountered by explicit experience repositories during long-term interactions, while simultaneously mitigating the tension caused by parameter updates lagging behind environmental dynamics.

\begin{itemize}[leftmargin=*, leftmargin=1.5em, rightmargin=1em]
    \item \textbf{Experience Transfer:} Experience Transfer facilitates capability internalization by progressively decoupling agents from external retrieval reliance. ~\citet{Wu2025EvolveRSL} employs offline distillation to abstract complex trajectories into structured experience for inference guidance, then uses these experiences to generate high-quality trajectories for policy optimization. By transferring knowledge from explicit experience pools into model parameters via gradient updates, this approach eliminates dependence on external retrieval systems.~\citet{liu2026exploratorymemoryaugmentedllmagent, Ouyang2025ReasoningBankSA} maintain an explicit experience replay buffer preserving high-value exploration trajectories. Through a hybrid On-Policy and Off-Policy update strategy, this framework leverages explicit memory for immediate exploration while encoding successful experiences into network parameters via offline updates, ensuring agents sustain optimal performance through internalized "intuition" without external support.
\end{itemize}

\section{Extended Discussion on Multimodal Memory Mechanisms}\label{Extended Discussion on Multimodal Memory Mechanisms}
Multimodal memory extends the conventional paradigm of memory centered on text to encompass heterogeneous modalities of perception, principally involving signals of a visual and auditory nature~\citep{Long2025SeeingLR}. The inputs for multimodal memory comprise text, audio, and images, with video constituting a composite input that integrates both auditory and visual components~\citep{Yin2025VideoARMAR}. As LLM agents begin to operate within hybrid environments that necessitate the joint execution of linguistic reasoning and perception across multiple dimensions, such as embodied navigation and interactive web browsing, mechanisms for memory are required to facilitate the capture of the rich dependencies across modalities that arise within interactions situated in the real world~\citep{Wang2026ExploreWL, Chen2026TowardsML, Yan2026M2DA}. In what follows, we summarize the principal body of work on multimodal memory and delineate the distinctive challenges that differentiate it from memory of a purely textual nature.
\subsection{Current Approaches.}\label{Current Approaches.}
Research on mechanisms for multimodal memory within LLM agents remains predominantly confined to the stage of Storage, with comparatively limited work addressing the stages of Reflection and Experience. We discuss existing methodologies along two salient dimensions: Multimodal Representation and Multimodal Retrieval.

\noindent\textbf{Multimodal Representation.}\label{Multimodal Representation.}
Unified semantic representation centered on text currently constitutes the predominant paradigm within mechanisms of multimodal memory for LLM agents~\citep{Liu2025MemVerseMM, Chen2025TeleMemBL}. The central premise of this approach resides in the utilization of pretrained Multimodal Large Language Models (MLLMs) for the textualization of all modalities; nevertheless, such a process incurs the loss of information inherent to the original modalities that resists adequate articulation within text. To mitigate the loss of visual information introduced through textualization, recent work has begun to construct dense representations of the original modalities, which are subsequently stored in juxtaposition with their corresponding summaries in text~\citep{Wen2026EventMemAgentHE, Bo2025AgenticLW}. Through indexing across multiple pathways and retrieval of an adaptive nature, these approaches seek to compensate for the attendant loss of information. Furthermore, embedding within a shared space is frequently invoked as one of the instruments leveraged by mechanisms of multimodal memory~\citep{He2024MALMMML}. Taken as a whole, existing research negotiates a trade-off between operations at the semantic level and fidelity to the original modalities, progressively transitioning from paradigms of pure textualization toward hybrid representations characterized by the juxtaposition of multiple modalities.

\noindent\textbf{Multimodal Retrieval.}\label{Multimodal Retrieval.}
Hybrid representations juxtapose multiple modalities, so queries and targets frequently cross modality boundaries, and different queries vary substantially in how much they demand semantic abstraction versus perceptual fidelity. To address this, three retrieval strategies have emerged. Parallel fusion retrieves along semantic, lexical, and visual channels simultaneously, then integrates the complementary signals through rank fusion to improve recall robustness~\citep{Feng2026M2AMM, Liu2026OmniSimpleMemAD}. Hierarchical retrieval cascades recall across levels of abstraction, first filtering coarsely over semantic summaries and then escalating to the original modalities for precise reconstruction~\citep{Lin2025HippoMMHM}. Agent-driven retrieval, by contrast, delegates the process to the agent itself, which selects modality-specific queries across multiple turns according to its current intent, until the acquired information is deemed sufficient~\citep{Yeo2025WorldMMDM}.

\subsection{Unique Challenges.}\label{Unique Challenges.}
In contrast to mechanisms of memory centered on text, multimodal memory confronts a series of distinctive challenges that arise from the heterogeneity across modalities. We elaborate this discussion principally along three dimensions, addressing Multimodal Alignment, Temporal Consistency, and subsequently Consolidation and Forgetting.

\noindent\textbf{Multimodal Alignment.}\label{Multimodal Alignment.}
Multimodal memory necessitates the binding of signals of a visual, auditory, and textual nature onto a unified semantic unit, a process considerably more arduous than alignment confined to a single modality. The granularity of signals across distinct modalities exhibits a pronounced asymmetry, the density of semantic content differs by a substantial margin, and the representations of a unified entity across disparate modalities may diverge entirely~\citep{Cai2025OnTV, Wang2025DeepMELAM, Yu2026ModalityGS}. Erroneous alignment not only precludes the recall of memory in its complete form, but may further introduce signals of an irrelevant character that impede the process of reasoning~\citep{Lu2026MMAMM}.

\noindent\textbf{Temporal Consistency.}\label{Temporal Consistency.}
Multimodal memory unfolds along a continuous axis of time rather than in terms of discrete tokens as with text, which renders the organization of length and granularity for units of memory a consideration of significant importance~\citep{Yang2025MemorySL, Lian2026FromVT}. Furthermore, the boundaries of events frequently fail to align with physical time: a single action may traverse visual variations spanning multiple frames as well as segments of speech of extended duration~\citep{Wang2024VideoAgentLV}. This state of affairs imposes upon mechanisms of memory the requirement for designs that incorporate adaptive segmentation, indexing across multiple scales, and consolidation of a hierarchical nature.

\noindent\textbf{Consolidation and Forgetting.}\label{Consolidation and Forgetting.}
In contrast to memory of a purely textual nature, wherein the processes of forgetting and consolidation may be executed within a unified semantic space, multimodal memory confronts a series of difficulties that are considerably more distinctive. First, the similarity across disparate modalities cannot be assessed through a unified metric~\citep{Wen2025QuantifyingCM, Wei2025MitigatingIA}. Second, the temporal validity associated with distinct modalities diverges by a pronounced margin: textual preferences may retain efficacy over extended durations, whereas the state of a scene may forfeit validity immediately following a transformation of the environment, rendering the adaptation of functions for decay a matter of considerable difficulty~\citep{Alqithami2025ForgetfulBF}. Finally, the process of consolidation may incur the loss of perceptual details inherent to signals of a non-textual nature~\citep{Lian2026FromVT}.

Taken as a whole, existing research on multimodal memory is concentrated predominantly within the stage of Storage, whereas work pertaining to the stages of Reflection and Experience remains exceedingly scarce. This distribution indicates that multimodal memory resides at an early phase of its development. Future research will likely address, at the level of Reflection, issues such as errors and hallucinations that manifest across modalities, and will further explore, within the stage of Experience, the extraction of policy priors that are invariant to modality from clusters of trajectories of a multimodal nature~\citep{Wang2025MitigatingMH, Allard2026ExperientialRL, Lei2025RoboMemoryAB}.

\section{Datasets and Benchmarks}\label{Datasets}

Recently, the research community has developed various datasets to evaluate the consistency over the long term and the capacity for self evolution of LLM agents within dynamic environments. However, existing benchmarks still primarily assess the storage and retrieval of static data, which results in a lack of evaluation for other critical capabilities of memory within scenarios of dynamic interaction. Following our proposed path of evolution (\S\ref{Evolutionary Path}), we categorize these benchmarks into stages of storage, reflection, and experience according to their primary areas of focus. Detailed information for these benchmarks is provided in Table~\ref{Table.1}.

\noindent\textbf{Storage Stage.} The storage stage serves as the cornerstone for mechanisms of memory, primarily evaluating the capacity of LLM agents for the accurate storage and retrieval of information over long distances across various scenarios, tasks, and modalities.




\begin{itemize}[leftmargin=*, leftmargin=1.5em, rightmargin=1em]
    \item \textbf{Extreme Context:} Extreme context types focus on probing the physical limits of memory in LLM agents, specifically the capacity for extracting and processing minute facts within massive volumes of distracting information. For instance, these benchmarks define the actual effective window of the model through the retrieval of multiple needles~\cite{Hsieh2024RULERWT}, assess the capabilities of agents by embedding reasoning tasks within backgrounds of a million words~\cite{Kuratov2024BABILongTT} and assessing the reliability of memory within a long context~\cite{Yen2024HELMETHT}, or extend these challenges to the domain of vision~\cite{Wang2024MultimodalNI}. The core of this area is the evaluation of the authentic capacity for memory in the model.
    \item \textbf{Interactive Consistency:} Research regarding the category of Interactive Consistency is based on interaction across sessions, which requires LLM agents to maintain memory with consistency throughout such interactions. Examples include the provision of frameworks for coherent dialogue at the scale of ten million words~\cite{Tavakoli2025BeyondAM}, the direct evaluation of the update of knowledge and the capacity for rejection during continuous interaction~\cite{Maharana2024EvaluatingVL}, and the detection of how consistency and accuracy for personas are maintained over histories of long duration~\citep{jia-etal-2025-evaluating, Zhong2023MemoryBankEL}. The core of this stage is the assessment of the capacity for memory with consistency over long distances.
    \item \textbf{Relational Fact:} Benchmarks of the relational fact category primarily evaluate the capacity of LLM agents for semantic association and reasoning across multiple hops. This involves testing the ability of the model for the integration of facts across documents and reasoning in multiple steps within the context of personal trivia~\citep{Zhang2025ExplicitVI, Yang2018HotpotQAAD}, while adjacent fact-checking evaluation frameworks have also involved knowledge recall and evidence integration during multi-step factual reasoning~\citep{lin2025fact, lin2026towards}. Furthermore, certain frameworks focus on emotional support and interactive scenarios to evaluate the model's capacity for memory recall across proactive and passive paradigms~\cite{He2024MADialBenchTR}.
\end{itemize}

\noindent\textbf{Reflection Stage.} The core of the Reflection stage is the evaluation of how agents transform raw trajectories into memory of high quality, with an emphasis on the denoising and fidelity of memory, the deep alignment with characteristics of users, and the support for perception within complex environments.




\begin{itemize}[leftmargin=*, leftmargin=1.5em, rightmargin=1em]
    \item \textbf{Error Correction:} Error correction primarily evaluates whether errors or hallucinations emerge during the lifecycle of the memory system. For instance, it involves testing operations for the search, editing, and matching of memory~\cite{Xia2025MinervaAP}, examining the presence of hallucinations during the stages of extraction or update~\cite{Chen2025HaluMemEH}.
    \item \textbf{Personalization:} Personalization focuses on the capacity for the extraction of deep personalization from the history of the agent, which includes the mining of latent information through reflection to identify implicit preferences~\citep{Jiang2025PersonaMemv2TP, Huang2025MemPALTM}, traits of users~\citep{Du2024PerLTQAAP, Zhao2025DoLR}, key information~\citep{Yuan2023PersonalizedLL, Li2025TowardMP}, and shared components~\citep{Tsaknakis2025DoLR, Kim2024SHARESM}.
    \item \textbf{Dynamic Reasoning:} Dynamic reasoning emphasizes the critical role of memory in reasoning across multiple steps and the perception of environments with high complexity. This involves the selective forgetting of memory~\cite{Hu2025EvaluatingMI}, backtracking on decisions~\cite{Wan2025StoryBenchAD}, scenarios in the real world~\citep{Deng2024OnTM, Miyai2025WebChoreArenaEW}, and the mechanisms for summarization and transition~\cite{Maharana2024EvaluatingVL}.
\end{itemize}



\noindent\textbf{Experience Stage.} The Experience Stage represents the pinnacle of the evolutionary path of memory mechanisms; at this phase, the focus shifts toward how LLM agents abstract general experience from fragmented trajectories within dynamic environments to facilitate continuous evolution through practical application. While benchmarks for this stage are relatively scarce, they possess a strong empirical character:~\citet{Wu2024StreamBenchTB} and~\citet{Ai2025MemoryBenchAB} simulate environments for authentic deployment to evaluate the capacity of LLM agents for the extraction and internalization of experience within cycles of input and feedback; conversely,~\citet{Wei2025EvoMemoryBL} and~\citet{Zheng2025LifelongAgentBenchEL} emphasize the capacity for the transfer of experience, measuring levels of abstraction and generalization by assessing the transfer of acquired experience to a diverse range of other tasks.

\begin{table*}[!htbp]
\centering
\scriptsize
\setlength{\tabcolsep}{2pt}
\renewcommand{\arraystretch}{0.93}
\resizebox{\textwidth}{!}{ \begin{tabular}{L{1.8cm} >{\hspace{0.3cm}}L{2.2cm} L{2.5cm} C{0.85cm} L{9.2cm}}
\toprule
\textbf{Stage} & \textbf{Dataset} & \textbf{Reference} & \textbf{Size} & \textbf{Description} \\
\midrule

\multirow{14}{1.8cm}[-6.5em]{Storage Stage Benchmark} 
& LongBench & ~\citet{Bai2023LongBenchAB} & 4.7k & Evaluate faithful memory preservation and retrieval by performing information extraction and reasoning across multiple tasks with sequences up to 32k tokens. \\
\cmidrule(l){2-5}
& LongBenchv2 & ~\citet{Bai2024LongBenchVT} & 503 & Answer complex multiple-choice questions through the processing of extremely long sequences with lengths between 8k and 2M words for the purpose of evaluating the capacity of memory and the precision of reasoning. \\
\cmidrule(l){2-5}
& RULER & ~\citet{Hsieh2024RULERWT} & Scalable & Evaluate the effectiveness of retrieval and synthesis within long contexts through tasks such as multi-needle extraction or multi-hop reasoning to identify true memory capacity. \\
\cmidrule(l){2-5}
& MMNeedle & ~\citet{Wang2024MultimodalNI} & 280k & Identify a target sub image within a massive collection of images through the analysis of textual descriptions and visual contents for the purpose of measuring the limits of multimodal retrieval. \\
\cmidrule(l){2-5}
& HotpotQA & ~\citet{Yang2018HotpotQAAD} & 113k & Answer questions by performing reasoning across multiple hops over information scattered in diverse documents based on Wikipedia to provide accurate results and supporting facts. \\    
\cmidrule(l){2-5}
& MemoryBank & ~\citet{Zhong2023MemoryBankEL} & 194 & Answer questions by recalling pertinent information and summarizing user traits across a history of interactions spanning ten days to evaluate the precision of retrieval and the maintenance of user portraits for long-term dialogues. \\
\cmidrule(l){2-5}
& BABILong & ~\citet{Kuratov2024BABILongTT} & Scalable & Answer questions by performing reasoning on facts scattered across extremely long documents of natural language to test the limits of memory and retrieval for contexts with length up to one million tokens. \\
\cmidrule(l){2-5}
& DialSim & ~\citet{Kim2024DialSimAR} & 1.3k & Evaluate the precision of retrieval for memory by answering spontaneous questions across sessions of dialogue involving multiple parties with long durations. \\
\cmidrule(l){2-5}
& LongMemEval & ~\citet{Maharana2024EvaluatingVL} & 500 & Answer questions through the extraction of information from histories of interactive chat with multiple sessions for the purpose of evaluating the performance of retrieval and reasoning across dependencies of long range. \\
\cmidrule(l){2-5}
& BEAM & ~\citet{Tavakoli2025BeyondAM} & 100 & Evaluate the capacity of memory and the precision of retrieval by answering questions based on coherent and topically diverse conversations with length up to ten million tokens. \\
\cmidrule(l){2-5}
& MPR & ~\citet{Zhang2025ExplicitVI} & 108k & Answer complex questions by conducting reasoning across multiple hops of factual information specific to a user within a framework of explicit or implicit memory for the purpose of evaluating the precision of retrieval. \\
\cmidrule(l){2-5}
& LOCCO & ~\citet{jia-etal-2025-evaluating} & 1.6k & Evaluate the persistence of memory and the retrieval of historical facts by analyzing chronological conversations across extended periods of time for the purpose of measuring information decay. \\
\cmidrule(l){2-5}
& MADial-Bench & ~\citet{He2024MADialBenchTR} & 160 & Evaluate the effectiveness of retrieval and recognition for historical events across multiple turns of interaction by simulating paradigms of passive and proactive recall for the purpose of providing emotional support. \\
\cmidrule(l){2-5}
& HELMET & ~\citet{Yen2024HELMETHT} & Scalable & Evaluate the effectiveness of models for long-context by performing information extraction and reasoning across seven diverse categories for sequences with lengths up to 128k tokens to provide a thorough assessment of memory capacity. \\

\bottomrule
\end{tabular}}
\caption{Representative datasets for benchmarking LLM agent memory mechanisms.}
\label{Table.1}
\end{table*}

\begin{table*}[!htbp]
\ContinuedFloat
\centering
\scriptsize
\setlength{\tabcolsep}{2pt}
\renewcommand{\arraystretch}{0.93}
\resizebox{\textwidth}{!}{ \begin{tabular}{L{1.8cm} >{\hspace{0.3cm}}L{2.2cm} L{2.5cm} C{0.85cm} L{9.2cm}}
\toprule
\textbf{Stage} & \textbf{Dataset} & \textbf{Reference} & \textbf{Size} & \textbf{Description} \\
\midrule

\multirow{15}{1.8cm}[-9em]{Reflection Stage Benchmark} 
& Minerva & ~\citet{Xia2025MinervaAP} & Scalable & Analyze the proficiency of LLMs in utilizing and manipulating context memory through a programmable framework of atomic and composite tasks for the purpose of pinpointing specific functional deficiencies and providing actionable insights. \\
\cmidrule(l){2-5}
& HaluMem & ~\citet{Chen2025HaluMemEH} & 3.5k & Evaluate the fidelity of memory by quantifying the occurrence of fabrication and omission during the stages of storage and retrieval across dialogues of multiple turns. \\
\cmidrule(l){2-5}
& MABench & ~\citet{Hu2025EvaluatingMI} & 2.1k & Evaluate the competencies of accurate retrieval and learning at test time across sequences of incremental interactions with multiple turns. \\
\cmidrule(l){2-5}
& PRM & ~\citet{Yuan2023PersonalizedLL} & 700 & Evaluate the capability of personalized assistants to maintain a dynamic memory bank by preserving evolving user knowledge and experiences across long term dialogues for the purpose of generating tailored responses. \\
\cmidrule(l){2-5}
& PersonMemv2 & ~\citet{Jiang2025PersonaMemv2TP} & 1k & Generate personalized responses through the extraction of implicit personas from interactions with long context and thousands of preferences of users to evaluate the adaptation of agents. \\
\cmidrule(l){2-5}
& LoCoMo & ~\citet{Maharana2024EvaluatingVL} & 50 & Evaluate the reliability of memory by executing question answering and event summarization across sequences of conversation with lengths of up to nine thousand tokens spanning thirty-five sessions. \\
\cmidrule(l){2-5}
& WebChoreArena & ~\citet{Miyai2025WebChoreArenaEW} & 451 & Analyze the performance of memory for information retrieval and complex aggregation by performing multiple steps of navigation and reasoning across hundreds of web pages. \\
\cmidrule(l){2-5}
& MT-Mind2Web & ~\citet{Deng2024OnTM} & 720 & Evaluate the performance of conversational web navigation via multiturn instruction following across sequential interactions with both users and environment for the purpose of managing context dependency and limited memory space. \\
\cmidrule(l){2-5}
& StoryBench & ~\citet{Wan2025StoryBenchAD} & 80 & Evaluate the capacity for reflection and sequential reasoning by navigating hierarchical decision trees within interactive fiction games to trace back and revise earlier choices independently across multiple turns. \\
\cmidrule(l){2-5}
& PerLTQA & ~\citet{Du2024PerLTQAAP} & 8.6k & Answer personalized questions by retrieving and synthesizing semantic and episodic information from a memory bank of thirty characters to evaluate the accuracy of retrieval for memory. \\
\cmidrule(l){2-5}
& ImplexConv & ~\citet{Li2025TowardMP} & 2.5k & Evaluate implicit reasoning in personalized dialogues by retrieving and synthesizing subtle or semantically distant information from history encompassing 100 sessions to test the efficiency of hierarchical memory structures. \\
\cmidrule(l){2-5}
& Share & ~\citet{Kim2024SHARESM} & 3.2k & Improve the quality of interactions across long durations by extracting persona data and memories of shared history from scripts of movies to sustain a consistent relationship between two individuals. \\
\cmidrule(l){2-5}
& Mem-PAL & ~\citet{Huang2025MemPALTM} & 100 & Evaluate the capability of personalization for assistants oriented toward service by identifying subjective traits and preferences of users from histories of dialogue and behavioral logs across multiple sessions for the purpose of generating tailored responses. \\
\cmidrule(l){2-5}
& PrefEval & ~\citet{Zhao2025DoLR} & 3k & Quantify the robustness of proactive preference following by evaluating the ability of models to infer and satisfy explicit or implicit user traits amidst long context distractions for sequences with length up to 100k tokens. \\
\cmidrule(l){2-5}
& LIDB & ~\citet{Tsaknakis2025DoLR} & 210 & Discover the latent preferences of users and generate personalized responses through interactions across multiple turns within a framework of three agents for the purpose of evaluating the efficiency of elicitation and adaptation. \\

\midrule

\multirow{4}{1.8cm}[-1.5em]{Experience Stage Benchmark} 
& StreamBench & ~\citet{Wu2024StreamBenchTB} & 9.7k & Evaluate the capacity for continuous improvement and online learning via iterative feedback processing across diverse task streams to measure the adaptation of agents after deployment. \\
\cmidrule(l){2-5}
& MemoryBench & ~\citet{Ai2025MemoryBenchAB} & 20k & Evaluate the capacity for continual learning of Large Language Model systems by simulating the accumulation of feedback from users across multiple domains to measure the effectiveness of procedural memory. \\
\cmidrule(l){2-5}
& Evo-Memory & ~\citet{Wei2025EvoMemoryBL} & 3.7k & Evaluate the capacity for learning at test time and the evolution of memory by processing continuous streams of tasks for the purpose of reuse of experience across diverse scenarios. \\
\cmidrule(l){2-5}
& LABench & ~\citet{Zheng2025LifelongAgentBenchEL} & 1.4k & Evaluate the lifelong learning ability and transfer of knowledge across sequences of interdependent tasks in dynamic environments for the purpose of measuring the acquisition and retention of skills. \\

\bottomrule
\end{tabular}}
\caption[]{Representative datasets for benchmarking LLM agent memory mechanisms (continued).}
\end{table*}

\end{document}